\documentclass[sn-aps]{sn-jnl}

\usepackage{graphicx}%
\usepackage{multirow}%
\usepackage{amsmath,amssymb,amsfonts}%
\usepackage{amsthm}%
\usepackage{mathrsfs}%
\usepackage[title]{appendix}%
\usepackage{xcolor}%
\usepackage{textcomp}%
\usepackage{manyfoot}%
\usepackage{booktabs}%
\usepackage{algorithm}%
\usepackage{algorithmicx}%
\usepackage{algpseudocode}%
\usepackage{listings}%

\usepackage[caption=false,font=footnotesize]{subfig}

\newif\ifpreprint
\preprintfalse
\preprinttrue

\jyear{2023}%

\begin{document}

\title[From Private to Public]{From Private to Public: Benchmarking GANs in the Context of Private Time Series Classification}

\author*[1,2]{\fnm{Dominique} \sur{Mercier}}\email{dominique.mercier@dfki.de}
\author[1,2]{\fnm{Andreas} \sur{Dengel}}\email{andreas.dengel@dfki.de}
\author[1]{\fnm{Sheraz} \sur{Ahmed}}\email{sheraz.ahmed@dfki.de}

\affil[1]{\orgdiv{Smart Data \& Knowledge Services}, \orgname{DFKI}, \orgaddress{\street{Trippstadter Str. 122}, \city{Kaiserslautern}, \postcode{67663}, \state{RLP}, \country{Germany}}}
\affil[2]{\orgdiv{Computer Science}, \orgname{RPTU}, \orgaddress{\street{Erwin-Schrödinger-Straße 52}, \city{Kaiserslautern}, \postcode{67663}, \state{RLP}, \country{Germany}}}

\abstract{Deep learning has proven to be successful in various domains and for different tasks. However, when it comes to private data several restrictions are making it difficult to use deep learning approaches in these application fields. Recent approaches try to generate data privately instead of applying a privacy-preserving mechanism directly, on top of the classifier. The solution is to create public data from private data in a manner that preserves the privacy of the data. In this work, two very prominent GAN-based architectures were evaluated in the context of private time series classification. In contrast to previous work, mostly limited to the image domain, the scope of this benchmark was the time series domain. The experiments show that especially GSWGAN performs well across a variety of public datasets outperforming the competitor DPWGAN. An analysis of the generated datasets further validates the superiority of GSWGAN in the context of time series generation.}


\maketitle

\section{Introduction}
\label{sec:introduction}
Machine learning has proven to be very successful recently. The success of a wide variety of algorithms in the field of artificial intelligence has been accompanied by a broadening of the application horizon. However, artificial intelligence is limited in several areas of our lives. One reason for this is the users' trust in the algorithms. While the advancements in the field of explicable artificial intelligence (XAI) have marginally overcome this challenge~\cite{arrieta2020explainable}, there are still plenty of open issues due to which applicability is limited. Data privacy plays a key role~\cite{zaeem2020effect} in several areas, including Medical equipment (Heart-lung machines, mechanical ventilation, robotic surgery, pacemaker devices, Dialysis machines) and related software, recreation, automotive, and aviation. Recently, with the development of the GDPR and the AI act, it is legally necessary to maintain data security. 

While this may sound simple in theory, it is a major challenge for neural networks. DNNs consist of millions of parameters, all of which can be used to leak data. Recently, several privacy-related attacks have been made on neural networks~\cite{liu2020privacy}. One well-known and effective method is the membership attack, which infers the membership of a data sample based on the model prediction. The easiest way to defend against such privacy-oriented attacks was anonymization~\cite{ding2010brief}, however, anonymization is no longer sufficient nowadays as DNNs can deanonymize data.

Neural networks are prone to privacy-oriented attacks, and it is quite challenging to avoid them. While the use of homomorphic encryption ensures absolute security and performance preservation, this method has proven to be very impractical in terms of additional time~\cite{sun2018private}. Another approach is the use of differential privacy during training. However, this can significantly reduce performance in time series analysis. A more recent approach deals with generative models~\cite{yoon2019time} that allow the production of synthetic data that cannot be distinguished from the original data. Using the synthetic data for the actual task ensures the security of the private data. This way, neural networks can be used in almost every field, such as medicine, finance, communication, and others. However, they need to satisfy privacy-preserving restrictions. To address the scarcity of benchmarking private generative models in the context of time series analysis in this area, this paper addresses the following points:

\begin{itemize}
    \item GAN-based private data generation to avoid privacy concerns
    \item Direct differential privacy vs. data generation
    \item Impact of generator and discriminator architecture
    \item Stopping criteria for generative model training
    \item Quantitative analysis of the quality of the generated data
\end{itemize}

\section{Related Work}
\label{sec:related}
The privacy-preserving domain consists of a wide variety of approaches that protect both the models and the data. One of the first protection and simplest approaches is anonymization~\cite{ding2010brief}. Anonymization involves modifying the data by deleting or changing properties. However, this method is no longer the gold standard, since so-called deanonymization methods can reconstruct the data. Another defense mechanism called differential privacy addresses learning, involving the influence of noise to hide private information~\cite{abadi2016deep}. The algorithm tries to counteract the so-called membership attack. During a membership attack, targeted inference queries are used to reconstruct the dataset using the gradients of the model~\cite{shokri2017membership}. However, the noise used in DP results in the model weights not accurately representing the training data and thus prevents this. Another approach that results in shallow gradients and is often used in collaborative learning is called federated learning. In FL, models are trained locally at the client side and their gradients are then sent to a globally averaged model~\cite{yang2019federated}. A well-known approach related to FL is the PATE framework, which is considered almost the standard~\cite{papernot2018scalable}. However, all of the above methods lead to a performance loss, as the model is limited in its learning process. To circumvent this, one can work with encryption, e.g., homomorphic encryption is one of the most promising techniques~\cite{sun2018private}. However, the use of HE is very impractical due to the extremely high computation times.

A method that was not directly developed for the privacy-preserving domain is the adversarial neural network. These networks are used to produce synthetic data. GANs were developed to circumvent data scarcity~\cite{yoon2019time}. Over time, various versions and improvements of this architecture have been developed. In addition, the GANs have been adapted to work conditionally, allowing to address the inequality of classes in a dataset. One of the most promising architectures is the Wasserstein GAN, which has been shown in the literature to converge better and produce higher quality results~\cite{arjovsky2017wasserstein} compared to the original GAN implementation. GANs in particular are interested in connection with data security. Methods like DP-GAN combine GANs and privacy-preserving methods to use the networks to safely generate data~\cite{xu2018dp}. This data resembles the original data but can be used without restrictions because it does not reveal any private properties. Different variations of the architectures and combinations have been published in this field. An improvement of the DP-GAN is the DPWGAN~\cite{xie2018differentially} where a Wasserstein GAN is used~\cite{xie2018differentially}. PATE-GAN describes the combination of GANs with the PATE framework, which can also be used in a federated setup~\cite{jordon2018pate}. Worth mentioning here is FED-GAN, which describes the use of GANs in a federated scenario. One of the latest approaches is GSWGAN~\cite{chen2020gs}, where differential privacy is replaced by gradient sanitization~\cite{chen2020gs} which leads to better performance and convergence according to the authors.

\section{Evaluated Method}
\label{sec:approach}
This section provides insights into the GAN-based methods used to evaluate the potential of privacy preservation in the context of time series classification. In particular, DPWGAN~\cite{xie2018differentially} and Gradient-sanitized Wasserstein GAN~\cite{chen2020gs} were used. Both of these methods share the same architecture, however, they have different privacy-preserving mechanisms. The main motivation to choose Wasserstein-based GANs~\cite{arjovsky2017wasserstein} for both DPWGAN~\cite{xie2018differentially} and GSWGAN~\cite{chen2020gs} is that the traditional GANs suffer from a vanishing gradient problem. However, the Wasserstein distance resolves this problem and provides a better learning behavior using linear gradients. 

The first method to generate privacy is the DPWGAN~\cite{xie2018differentially}, which is a Wasserstein GAN (WGAN) where differential privacy is applied. To ensure privacy, noise is added to the gradients and additionally, the gradients of the discriminator are clipped. The amount of noise is crucial for the privacy of the model. According to Xie et al., in addition to the noise, weight clipping is essential to obtain a good model performance. It is important to mention that the noise added to ensure privacy and weight clipping are applied to the discriminator and affect the generator in the backpropagation.

The second method is the GSWGAN~\cite{chen2020gs}. The generator receives sanitized gradients from the discriminator. The idea is that it is not necessary to protect the discriminator from adversarial attacks, since it is discarded after training. An advantage of this method is that the discriminator is not limited in performance. Furthermore, the gradients of the discriminator are not clipped to a fixed value, so the loss of performance is lower.

\section{Datasets}
\label{sec:datasets}
The experiments below were conducted across multiple datasets of the UEA \& UCR repository~\cite{tsc2021datasets}. The datasets were selected in a way that covers multiple different aspects such as a different number of channels, classes, dataset size, and problem statement. Furthermore, the selection covers different domains to emphasize the broad applicability of the methods as well as to highlight limitations. This way, the selection of datasets presents a comprehensive set covering both easy and difficult datasets concerning the generation of data.

\begin{table}[!t]
\caption{Datasets related to critical infrastructures. Different characteristics such as the dataset size, length, feature number, and classes are covered by this selection.}
\label{tab:datasets}
\centering
\begin{tabular}{l|r|r|r|r|r}
\textbf{Domain \& Dataset} & \textbf{Train} & \textbf{Test} & \textbf{Steps} & \textbf{Chls.} & \textbf{Cls.} \\
\hline\hline
\textbf{Transport Systems} & & & & & \\
AsphaltPavementType     & 1,055   & 1,056   & 1,543   & 1   & 3 \\
AsphaltRegularity       & 751     & 751     & 4,201   & 1   & 2 \\
\hline
\textbf{Communications} & & & & & \\
CharacterTrajectories   & 1,422   & 1,436   & 182     & 3   & 20 \\
HandOutlines            & 1,000   & 370     & 2,709   & 1   & 2 \\
UWaveGestureLibraryAll  & 896     & 3,582   & 945     & 1   & 8 \\
Wafer                   & 1,000   & 6,164   & 152     & 1   & 2 \\
\hline
\textbf{Pulbic Health} & & & & & \\
ECG5000                 & 500     & 4,500   & 140     & 1   & 5 \\
FaceDetection           & 5,890   & 3,524   & 62      & 144 & 2 \\
\hline
\textbf{Critical Manufacturing} & & & & & \\
FordA                   & 3,601   & 1,320   & 500     & 1   & 2 \\
\end{tabular}
\end{table}

\section{Experiments \& Results}
\label{sec:experiments}
To evaluate the performance of selected GANs private data is generated, and a classification task is performed. To do so, InceptionTime~\cite{fawaz2020inceptiontime} was used as a classification network, as it produces state-of-the-art performances for the datasets. During preprocessing, the datasets were standardized and normalized to range between zero and one for the sigmoid function of the generator. As a baseline, an InceptionTime network was trained for each dataset using a learning rate scheduler and Adam optimizer. These models are referred to as private models as they have trained directly on real UEA \& UCR~\cite{tsc2021datasets} datasets and should not be shared.

\subsection{Experiment 1: Accuracy Comparison of DP, DPWGAN, and GSWGAN}
\begin{table}[!t]
\caption{Accuracy Comparison: Shows the weighted f1-scores for classifiers trained on private or public data. Private (d-) and public (d+) correspond to the privacy of the evaluation dataset. The model (InceptionTime) was trained on private (m-) and public (m+) data. A noise multiplier of 0.5 was used for all privacy-preserving approaches.} 
\label{tab:comparison}
\centering
\resizebox{\columnwidth}{!}{
\begin{tabular}{l||r|r||r|r|r|r|r|r}
\textbf{Dataset} &
\textbf{Baseline} &
\textbf{DP}~\cite{abadi2016deep} &
\multicolumn{2}{c|}{\textbf{WGAN}~\cite{arjovsky2017wasserstein}} &
\multicolumn{2}{c|}{\textbf{DPWGAN}~\cite{xie2018differentially}} &
\multicolumn{2}{c}{\textbf{GSWGAN}~\cite{chen2020gs}} \\
& & & \textbf{m- d+} & \textbf{m+ d-} & \textbf{m- d+} & \textbf{m+ d-} & \textbf{m- d+} & \textbf{m+ d-} \\ 
\hline\hline
AsphaltPavementType & 
0.9082 & 0.8356 & 0.7845 & 0.6955 & 0.5734 & 0.3472 & 0.5730 & 0.7299 \\
\hline
AsphaltRegularity &
0.9973 & 0.9614 & 0.9374 & 0.9479 & 0.8161 & 0.4575 & 0.7652 & 0.8399 \\
\hline
CharacterTrajectories &
0.9965 & 0.8955 & 0.9716 & 0.9749 & 0.6006 & 0.5574 & 0.9772 & 0.9887 \\
\hline
ECG5000 &
0.9318 & 0.8955 & 0.8769 & 0.5329 & 0.8503 & 0.1906 & 0.9189 & 0.8811 \\
\hline
FaceDetection &
0.4944 & 0.5973 & 0.3324 & 0.3592 & 0.3333 & 0.3416 & 0.3460 & 0.3340 \\
\hline
FordA & 
0.9492 & 0.9417 & 0.9485 & 0.9250 & 0.4236 & 0.4690 & 0.9606 & 0.9220 \\
\hline
HandOutlines &
0.6540 & 0.5002 & 0.6763 & 0.5347 & 0.4573 & 0.7246 & 0.6720 & 0.5304 \\
\hline
UWaveGestureLibraryAll &
0.9032 & 0.4238 & 0.9264 & 0.8658 & 0.8566 & 0.7031 & 0.9401 & 0.8461 \\
\hline
Wafer &
0.9979 & 0.9521 & 0.9880 & 0.9769 & 0.9794 & 0.8581 & 0.9808 & 0.8501 \\
\hline\hline
\textbf{Average} &
0.8703 & 0.7781 & 0.8269 & 0.7570 & 0.6545 & 0.5166 & 0.7926 & 0.7691 \\
\end{tabular}
}
\end{table}

In this experiment, the capabilities of generative models and the direct application of privacy on the classifier are compared. In addition, the difference between public, differential privacy, and gradient-sanitized generative models is shown. Therefore, for each dataset first, a private classifier was trained as a baseline. Furthermore, a differentially private version of the classifier was trained. Next, WGAN~\cite{arjovsky2017wasserstein}, DPWGAN~\cite{xie2018differentially}, and GSWGAN~\cite{chen2020gs} were trained on the real data. This results in a private generator for the WGAN and public generators for the other approaches. Using the data generated by the generators, a classifier without privacy constraints was trained. Except for the WGAN generator, this results in a classifier that preserves privacy due to the privacy constraints involved during the training of the generator. Table~\ref{tab:comparison} shows the results for different approaches for both the classification of the private and the public-generated datasets. The models and the data are denoted with either a plus or a minus sign, highlighting the used data and the training conditions. Models with a minus (m-) correspond to the private baseline classifier and therefore cannot be used in public, whereas models with a plus (m+) are trained on generated data and can be used in public, except for the WGAN approach. Data with a plus (d+) corresponds to the generated test data, and data with a minus (d-) corresponds to the private test data. Throughout the rest of the paper, this denotation is used. In addition, InceptionTime was always used as a classifier, whereas the number of layers for the generative models varied based on the datasets.

Table~\ref{tab:comparison} shows that applying differential privacy on the classifier directly reduces the performance by about 10\% on average compared to the baseline without privacy. However, the accuracy drop strongly varies between the datasets, e.g., the accuracy for the UWaveGestureLibraryAll dataset has declined by 48\% whereas the FordA dataset has shown a decline of less than one percent. Overall, the DP approach converged for all datasets, but does not enable sharing of any data. In contrast to that, the DPWGAN~\cite{xie2018differentially} and GSWGAN~\cite{chen2020gs} allow the exchange of the data as it is created synthetically. The WGAN serves as a baseline public generator to compare the private generators against. The average performance of the WGAN was about 12\% lower when the classifier was trained on public data produced by the WGAN and tested on the private. Similarly, training the classifier on the private data and testing on the public data generated by the WGAN resulted in a 5\% performance loss. 

Intuitively, adding DP constraints to the WGAN resulted in lower performance. The DPWGAN~\cite{xie2018differentially} showed a drop of 36\% when tested on the private data and trained on the generated data. For most of the datasets, the model shows a significantly lower performance due to the additional privacy constraints that limit the discriminator and therefore the generative capabilities. GSWGAN~\cite{chen2020gs} has shown an average performance drop of 10\% compared to the baseline. GSWGAN~\cite{chen2020gs} is on par with the DP classifier but, in addition, enables further use of the generated data and classifier. Across all datasets, its performance was stable. The superior performance can be explained by the fact that privacy is only applied to the generator, as this is the only attackable part. It has to be mentioned that the FaceDetection dataset overall has shown a bad performance independent of the classifier and generative model.

\subsection{Experiment 2: Finding the Best Stopping Criteria for GSWGAN}
Training a generative model is a challenging task as these models can collapse or produce worse samples when trained for too many iterations. Usually, when it comes to privacy-preserving approaches these are trained for a fixed number of iterations based on the available privacy budget, however, using the complete available budget does not mandatory yield the best performance and is costly. The stopping criteria evaluated in this experiment are listed below:

\begin{itemize}
\item 
Training the GAN without any stopping criteria for a fixed number of iterations results in using the maximum privacy budget. However, this method requires a lot of time and can result in a mode collapse during the training.

\item 
The Frechet Inception Distance introduced by Heuse et al.~\cite{heusel2017gans} measures the distance of the samples in the latent space of a classifier model. This method initially was introduced for the image domain, but can be reformulated to work on a time series. Using this measurement provides evidence that the generated data is close to the original data.

\item 
The Inception Score proposed by Salmimans et al.~\cite{salimans2016improved} computes the similarity of the distribution based on the samples. Initially, this was proposed for the image domain, but it works in the time series domain too. The idea is to measure the quality of the samples based on their similarity. 

\item
Loss and accuracy are two well-known direct metrics. However, these do not provide any information about the quality of the generated data, but rather focus on the performance of the classifier attached to the GAN. The drawback of this method is that the data can look quite different.
\end{itemize}

For instance, for metrics that require an additional classifier, the trained InceptionTime of the baseline can be used. This does not lower the privacy, as the measurements are not back-propagated to the generative models. However, this means that additional time is required to create the baseline model. Although this requires additional time, the training of a classifier to evaluate the quality of the samples is faster than the training of the generative model for a larger number of epochs, and in addition, it provides insights into the upper bound for a classifier trained on the generated data.

In Table~\ref{tab:stopping} the results of GSWGAN~\cite{chen2020gs} on a set of selected datasets are presented. The results on the remaining datasets do not provide additional insights and are excluded. Furthermore, DPWGAN~\cite{xie2018differentially} was excluded, as the results in Table~\ref{tab:comparison} emphasize the use of GSWGAN~\cite{chen2020gs}. The rest of this work focuses on GSWGAN~\cite{chen2020gs} as it is the most promising concerning stability and performance. The results provide evidence that the FID score produces better results on average, leading to a higher quality of generated samples. The classifier performance using the FID showed that the average performance is 3\% better than using the accuracy. The worst result was achieved using the IS score, which showed a drop of 7\% compared to the FID. Training for a fixed number of iterations has shown the second worse results. This provides evidence that the FID score, which measures the similarity in the latent representation, is a good measurement to use as an early stopping criterion for the privacy-preserving training of GSWGAN~\cite{chen2020gs}. The largest gains were observed for the FaceDetection and the Wafer dataset. In the case of the FaceDetection dataset, the FID approach shows an increase of 15\% compared to excluding the stopping criteria. The performance on the Wafer dataset shows an increase of 10\% for the same case.

\begin{table}[!t]
\caption{Stopping criteria: F1-scores of classifiers trained on generated data of GSWGAN~\cite{chen2020gs} and tested on the private dataset (m+ d-). Limit of 50,000 iterations and patience of 2,500 iterations.} 
\label{tab:stopping}
\centering
\resizebox{\columnwidth}{!}{
\begin{tabular}{l|r|r|r|r|r}
\textbf{Dataset} & \textbf{Fixed iterations} & \textbf{FID~\cite{heusel2017gans}} & \textbf{IS~\cite{salimans2016improved}} & \textbf{Loss} & \textbf{Accuracy} \\
\hline\hline
CharacterTrajectories &
0.9887 & 0.9895 & 0.9840 & 0.9772 & 0.9826 \\
\hline
ECG5000 &
0.8811 & 0.8939 & 0.8845 & 0.8890 & 0.8890 \\
\hline
FaceDetection &
0.3340 & 0.4833 & 0.3356 & 0.4407 & 0.5013 \\
\hline
Wafer &
0.8501 & 0.9576 & 0.8328 & 0.8328 & 0.8206 \\
\hline\hline
\textbf{Average} &
0.7635 & 0.8311 & 0.7592 & 0.7850 & 0.7984 \\
\end{tabular}
}
\end{table}

\subsection{Experiment 3: Impact of Architecture on GSWGAN}
\begin{table}[!t]
\caption{Architecture: Shows the f1-score of InceptionTime classification using the GSWGAN~\cite{chen2020gs} approach. The IS score highlights closer data when using the convolutional approach.} 
\label{tab:arch}
\centering
\begin{tabular}{l|r|r|r|r|r|r}
\multirow{2}{*}{\textbf{Dataset}} & 
\multicolumn{3}{c|}{\textbf{GSWGAN-dense}} &
\multicolumn{3}{c}{\textbf{GSWGAN-conv}} \\
& \textbf{m- d+} & \textbf{m+ d-} & \textbf{IS~\cite{salimans2016improved}} &
\textbf{m- d+} & \textbf{m+ d-} & \textbf{IS~\cite{salimans2016improved}} \\
\hline\hline
CharacterTrajectories &
0.9807 & 0.9895 & 18.1667 & 0.9930 & 0.9867 & 19.2626 \\
\hline																		
ECG5000 &
0.9236 & 0.8939 & 1.8705 & 0.9443 & 0.5831 & 1.8722 \\
\hline
FaceDetection &
0.3379  & 0.4833  & 1.0206 & 0.4867  & 0.5851  & 1.0574 \\
\hline
Wafer &
0.9755 & 0.9576 & 1.2756 & 0.9874 & 0.9898 & 1.3491 \\
\hline\hline																		
\textbf{Average} &
0.8044 & 0.8311 & 5.5834 & 0.8528 & 0.7861 & 5.8853 \\
\end{tabular}
\end{table}

In this experiment, the difference between a dense and a convolutional setup for the generative models was tested on four datasets. Therefore, GSWGAN-dense was created using fully connected layers, while GSWGAN-conv uses convolutional and transposed convolutional layers. However, using convolutional layers requires defining different strides filter sizes, leading to a closer optimization of the model architecture as these parameters depend on the dataset parameters. The fully connected network does not require these parameters, making it easier to find a working architecture. Based on the previous findings, the networks are trained with early stopping based on the FID score. 

In Table~\ref{tab:arch} the results are shown. It is visible that the convolutional setup achieves a better performance across all datasets. This is further reflected by the IS scores, as the convolutional one is four times higher than the dense one. Concerning the performance, the convolutional setup showed an increase of 4\% on the private test data, with the largest increase of 10\% for the FaceDetection dataset. As the training of the GSWGAN~\cite{chen2020gs} requires a lot of time to converge, it is essential to find an architecture that works without a comprehensive grid search on the model architecture. In the experiment presented in Table~\ref{tab:arch}, no additional architecture search was required. However, for complex datasets such as the FordA, it is required to perform an architecture search. In the case of the FordA different numbers of layers, filters, and latent dimension sizes were evaluated. The experiments cover only a subset of interesting architectures, emphasizing on the most relevant differences. The setup only covers GSWGAN-conv as the previous analysis indicated that the convolutional setup achieves better results, and second the dense setups did not converge for the FordA dataset.

Table~\ref{tab:arch_grid} shows the subset of different convolutional setups. The f1-scores of the baseline InceptionTime classifier for this dataset was 0.9492 and the best-performing architecture achieved 0.9220. Therefore, with the correct architecture, the model dropped only 2.7\% in performance. However, the results provide evidence that with a worse selection of the architecture, the performance can drop to 0.4149. Furthermore, the results indicate that the increase in the z-dimension did not provide any improvement. However, the filter size and filter number play an important role. As a result, the architecture search can be expensive, especially as training a generative model is costly.

\begin{table}[!t]
\caption{Architecture Search: Shows the weighted f1-scores for InceptionTime classifier trained on private data (m-) and on public data (m+). GSWGAN~\cite{chen2020gs} was used to generate the data. The baseline performance was 0.9492.} 
\label{tab:arch_grid}
\centering
\resizebox{\columnwidth}{!}{
\begin{tabular}{r|r|r|r|r|r|r}
\multicolumn{3}{c|}{\textbf{Network parameters}} &
\multirow{2}{*}{\textbf{m- d+}} &
\multirow{2}{*}{\textbf{m+ d-}} &
\multirow{2}{*}{\textbf{FID~\cite{heusel2017gans}}} &
\multirow{2}{*}{\textbf{IS~\cite{salimans2016improved}}} \\
\textbf{z-dim} & \textbf{Number of Filters} & \textbf{Size} & & & & \\
\hline\hline
32 & 256-256-256-256 & 7-5-3                  & 0.7043  & 0.7210  & 16.5325  & 1.4347 \\ 
\textbf{32} & \textbf{512-256-128-128-64-64} & \textbf{7-5-5-3-3} & 
\textbf{0.9606}  & \textbf{0.9220}  & \textbf{1.3951}  & \textbf{1.7158}  \\
32 & 512-256-128-64 & 7-5-3                   & 0.7781  & 0.7804  & 16.9240 & 1.4585  \\

48 & 512-256-128-64 & 7-7-7                   & 0.7300  & 0.7125  & 15.0901 & 1.4678  \\
48 & 512-512-512-512 & 7-7-7                  & 0.9644  & 0.8194  & 4.0653  & 1.8208  \\

64 & 512-256-128-64 & 7-5-3                   & 0.7308  & 0.7473  & 16.3632 & 1.4303  \\
64 & 512-256-128-64 & 7-7-7                   & 0.7058  & 0.4149  & 15.7679 & 1.4339  \\
\end{tabular}
}
\end{table}

\subsection{Experiment 4: Impact of Noise Multiplier on Privacy-preserving Approaches}
As the goal of privacy-preserving machine learning is to achieve sufficient privacy using the budget, it is important to understand the impact of the budget on the outcome performance. The noise used in the DP learning process is one of the main parameters that affect privacy. The selection of the noise multiplier depends on multiple parameters such as the dataset size, model iterations during training, and desired privacy budget. In this experiment, the noise multiplier is increased from 0.25 to 2.0. The DP classifier was also included, although it is not possible to directly compare the performance, as the number of iterations differs. An important point is that increasing the noise multiplier increases privacy.

In Table~\ref{tab:noise} the results are shown for a subset of datasets. The DP classifier shows that there is a significant difference based on the datasets. Whereas the performance drop for the ECG500 dataset is less than 2\% for all multipliers, for most of the other datasets the model fails at a certain point. Especially for the CharacterTrajectories dataset, a noise multiplier of 0.5 was enough to decrease the performance by more than 40\%. The rest of the datasets validated the finding that the DP classifiers showed significant performance drops compared to the generative approaches. In comparison to that, the DPWGAN~\cite{xie2018differentially} and GSWGAN~\cite{chen2020gs} were able to produce datasets that resulted in converging classifiers across all datasets. The DPWGAN~\cite{xie2018differentially} showed a similar performance decrease with an increasing noise multiplier. However, the GSWGAN~\cite{chen2020gs} achieved much higher performances across all the datasets when compared to the other two approaches. Except for the FaceDetection dataset, GSWGAN~\cite{chen2020gs} performed better in every setup. In the case of the CharacterTrajectories dataset, the performance drop for the noise multiplier of 2.0 was 25\%. Up to a noise multiplier of 1.5, the performance of GSWGAN~\cite{chen2020gs} did not show significant performance drops. Overall, the noise multiplier experiments provide evidence that the GSWGAN~\cite{chen2020gs} is more robust to the noise added. This is the case as the discriminator does not suffer from privacy concerns and only the generator is affected by the noise. Therefore, it is possible to have a much stronger discriminator even with high noise values.

\begin{table}[!t]
\caption{\textbf{Impact Noise:} Shows the impact of the noise on the different architectures. The DP Baseline was trained on the private data, whereas accuracies of the GAN are computed using a classifier trained on the generated public datasets.} 
\label{tab:noise}
\centering
\resizebox{\columnwidth}{!}{
\begin{tabular}{l|l|r|r|r|r|r}
\multirow{2}{*}{\textbf{Dataset}} &
\multirow{2}{*}{\textbf{Method}} &
\multicolumn{5}{c}{\textbf{Noise}} \\
& & \textbf{0.25}  & \textbf{0.5}  & \textbf{1.0}  & \textbf{1.5}  & \textbf{2.0} \\
\hline\hline
\multirow{3}{*}{CharacterTrajectories} & DP~\cite{abadi2016deep} &
0.8114 & 0.3986 & 0.1824 & 0.0883 & 0.0682 \\
& DPWGAN~\cite{xie2018differentially} &
0.8092 & 0.7479 & 0.4765 & 0.3497 & 0.2727 \\
& GSWGAN~\cite{chen2020gs} &
0.9558 & 0.9895 & 0.9650 & 0.9041 & 0.6999 \\
\hline
\multirow{3}{*}{ECG500} & DP~\cite{abadi2016deep} &
0.8961 & 0.8949 & 0.8902 & 0.8824 & 0.8765 \\
& DPWGAN~\cite{xie2018differentially} &
0.2739 & 0.8545 & 0.4490 & 0.4302 & 0.0953 \\
& GSWGAN~\cite{chen2020gs} &
0.8753 & 0.8939 & 0.8944 & 0.8958 & 0.7364 \\
\hline
\multirow{3}{*}{FaceDetection} & DP~\cite{abadi2016deep} &
0.5964 & 0.5871 & 0.5601 & 0.3377 & 0.3353 \\
& DPWGAN~\cite{xie2018differentially} &
0.4671 & 0.3423 & 0.5054 & 0.5278 & 0.3510 \\
& GSWGAN~\cite{chen2020gs} &
0.4468 & 0.4833 & 0.5114 & 0.3744 & 0.3392 \\
\hline
\multirow{3}{*}{Wafer} & DP~\cite{abadi2016deep} &
0.9522 & 0.8412 & 0.8412 & 0.8412 & 0.8412 \\
& DPWGAN~\cite{xie2018differentially} &
0.8533 & 0.8412 & 0.0210 & 0.6507 & 0.8118 \\  
& GSWGAN~\cite{chen2020gs} &
0.9692 & 0.9576 & 0.8552 & 0.9602 & 0.7152 \\
\end{tabular}
}
\end{table}

\subsection{Experiment 5: T-SNE Visualization of Generated Data}
\begin{figure}[!t]
\centering
\subfloat[CharacterTrajectories: DPWGAN~\cite{xie2018differentially}]{
\includegraphics[width=.48\linewidth]{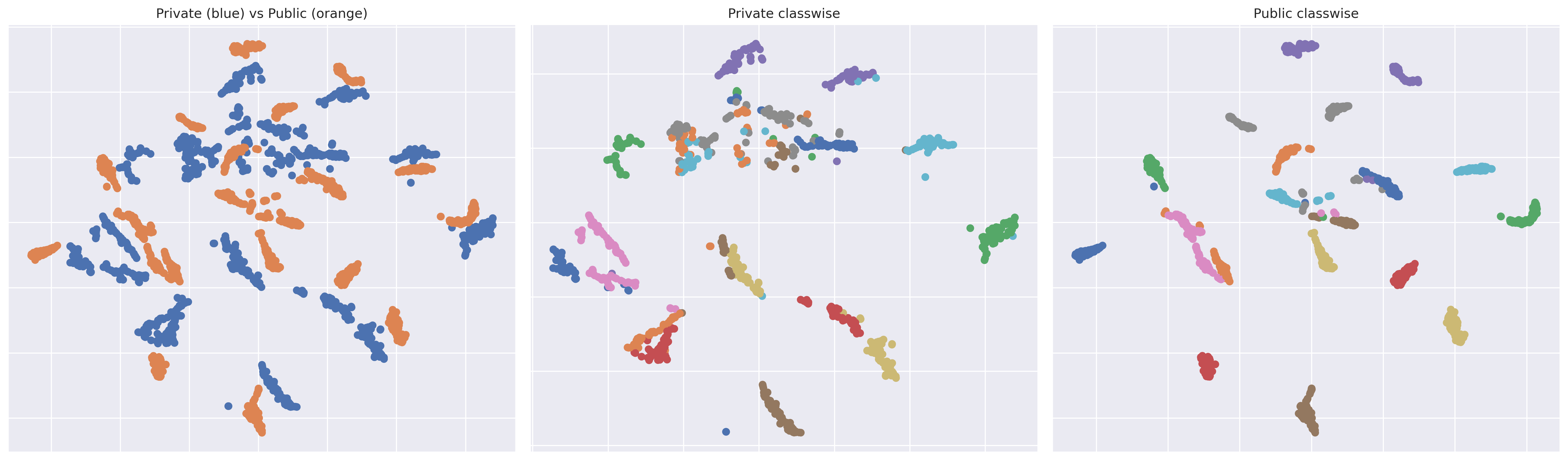}
\label{fig:char_tsne_dpgan}
}
\subfloat[CharacterTrajectories: GSWGAN~\cite{chen2020gs}]{
\includegraphics[width=.48\linewidth]{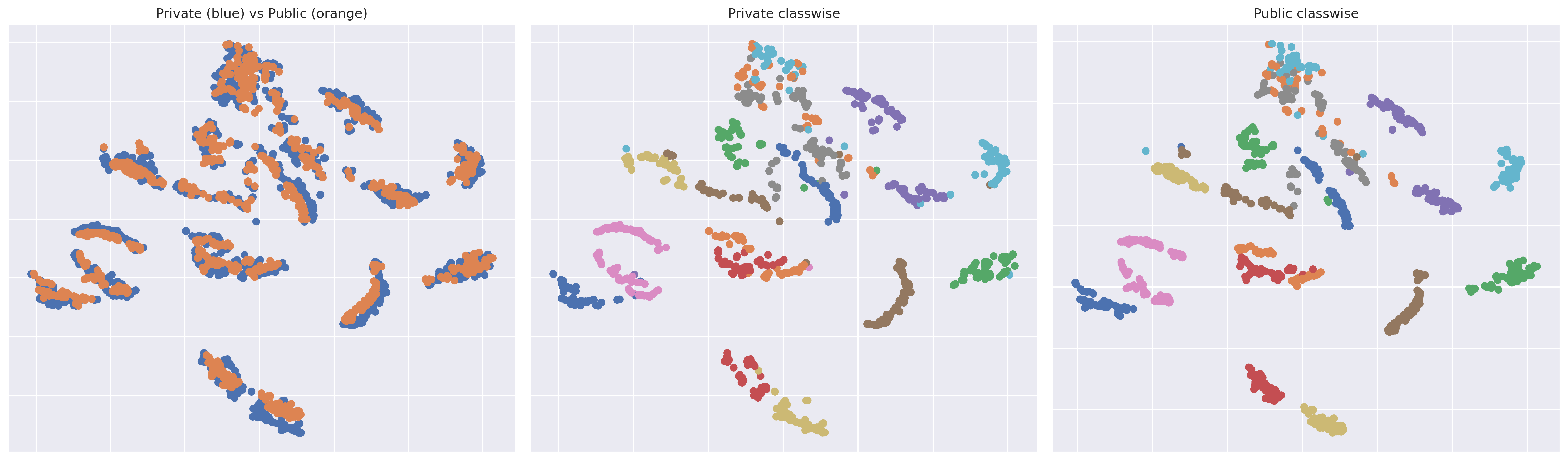}
\label{fig:char_tsne_gscwgan}
}
\hfil
\subfloat[ECG5000: DPWGAN~\cite{xie2018differentially}]{
\includegraphics[width=.48\linewidth]{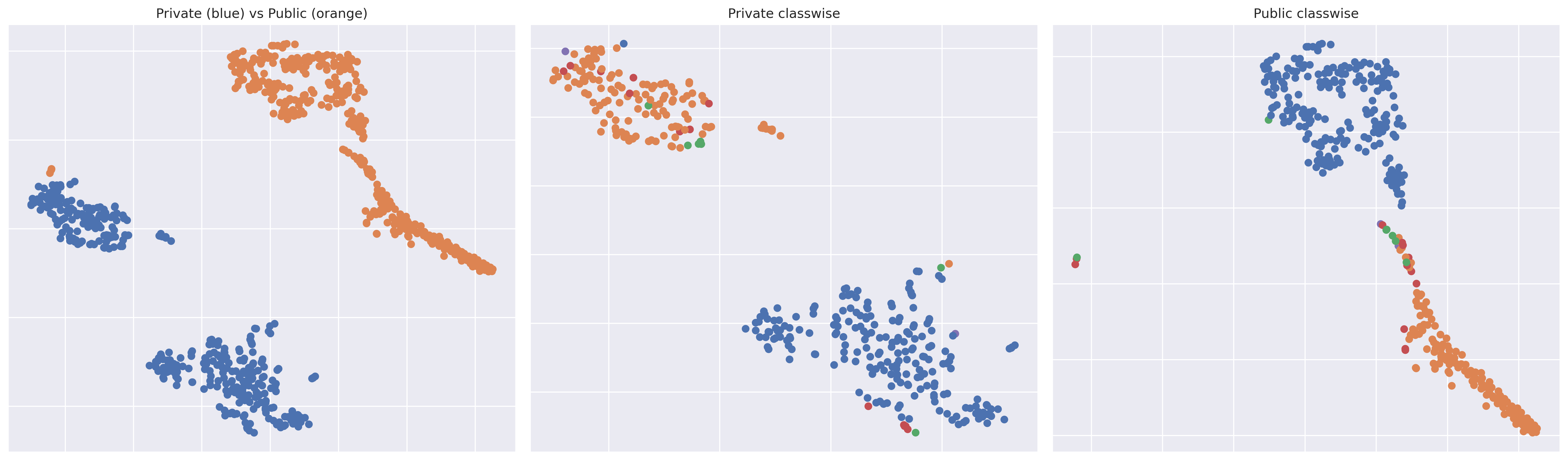}
\label{fig:ecg_tsne_dpwgan}
}
\subfloat[ECG5000: GSWGAN~\cite{chen2020gs}]{
\includegraphics[width=.48\linewidth]{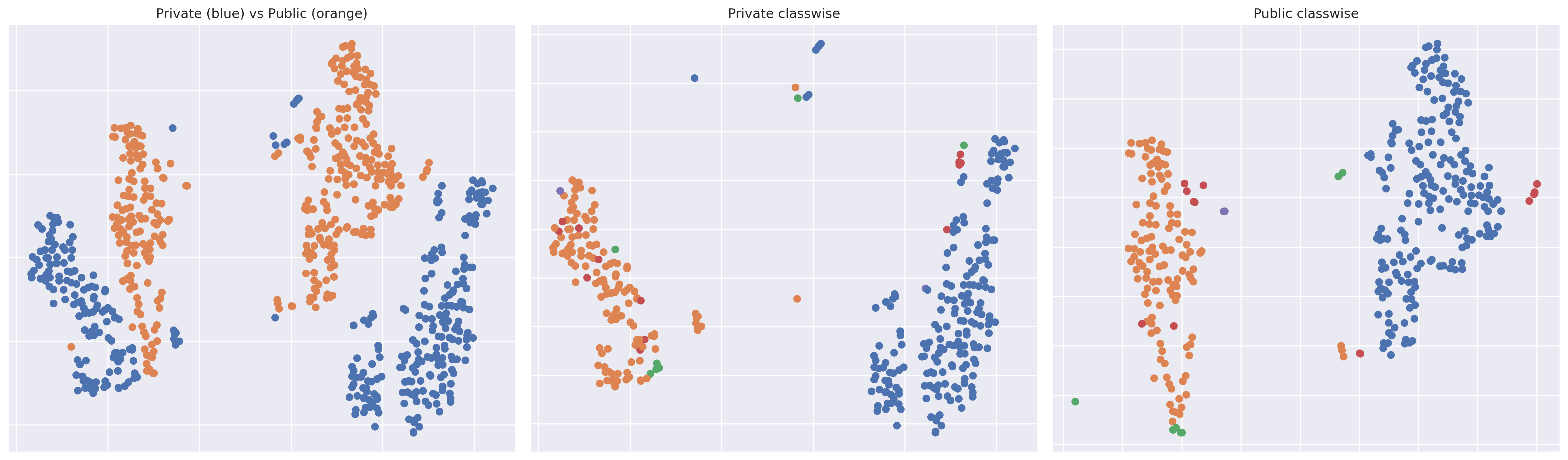}
\label{fig:ecg_tsne_gscwgan}
}
\hfil
\subfloat[FaceDetection: DPWGAN~\cite{xie2018differentially}]{
\includegraphics[width=.48\linewidth]{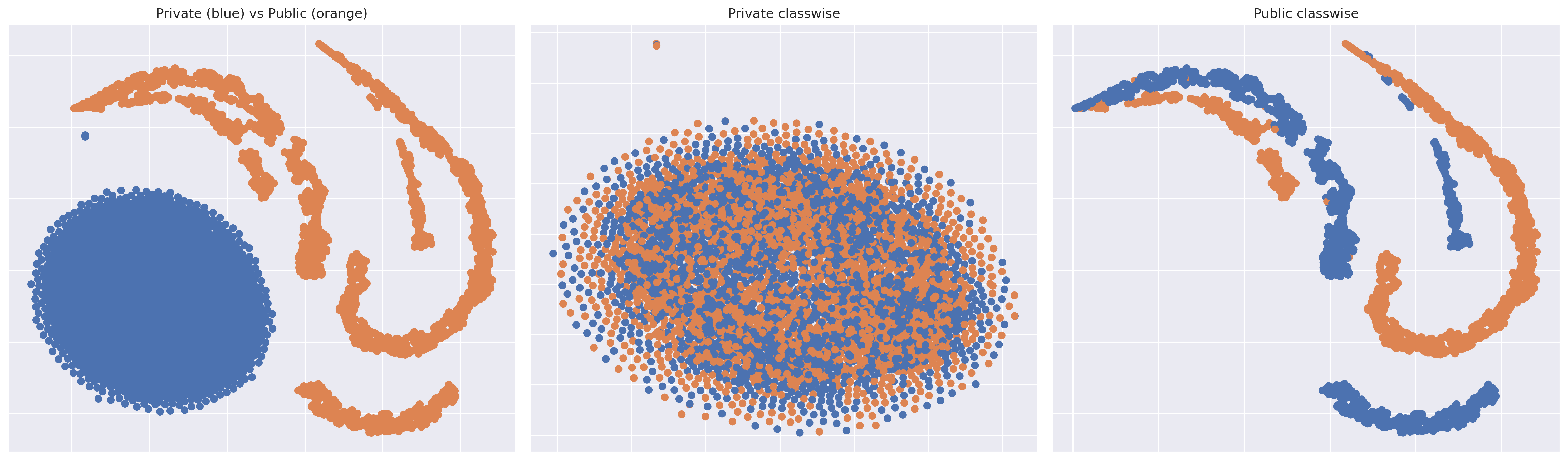}
\label{fig:face_tsne_dpwgan}
}
\subfloat[FaceDetection: GSWGAN~\cite{chen2020gs}]{
\includegraphics[width=.48\linewidth]{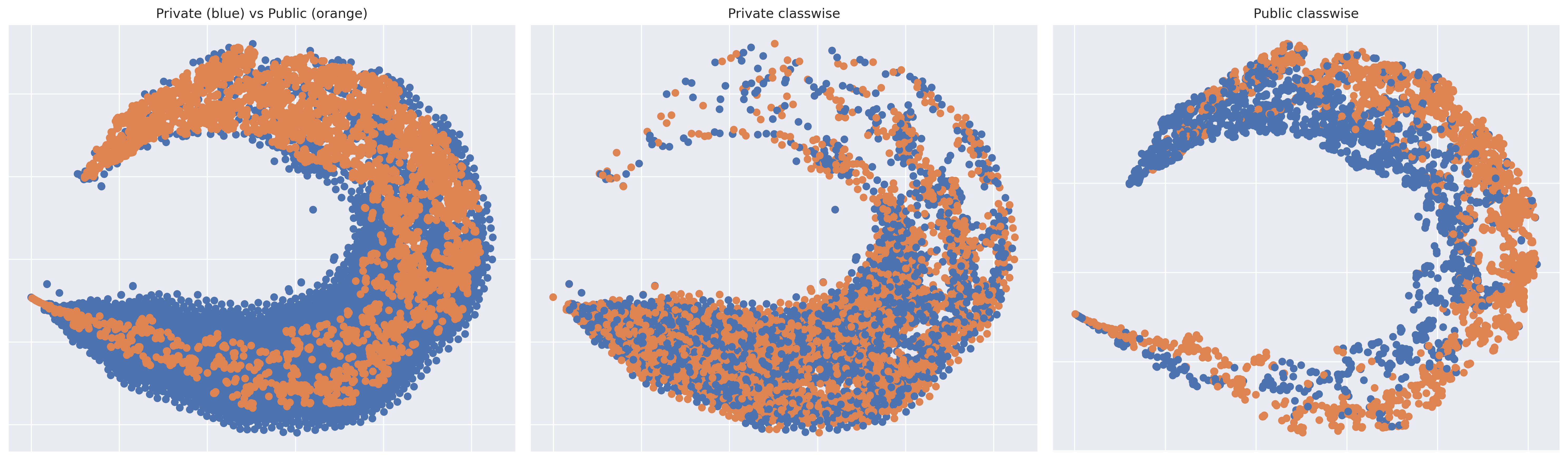}
\label{fig:face_tsne_gscwgan}
}
\hfil
\subfloat[Wafer: DPWGAN~\cite{xie2018differentially}]{
\includegraphics[width=.48\linewidth]{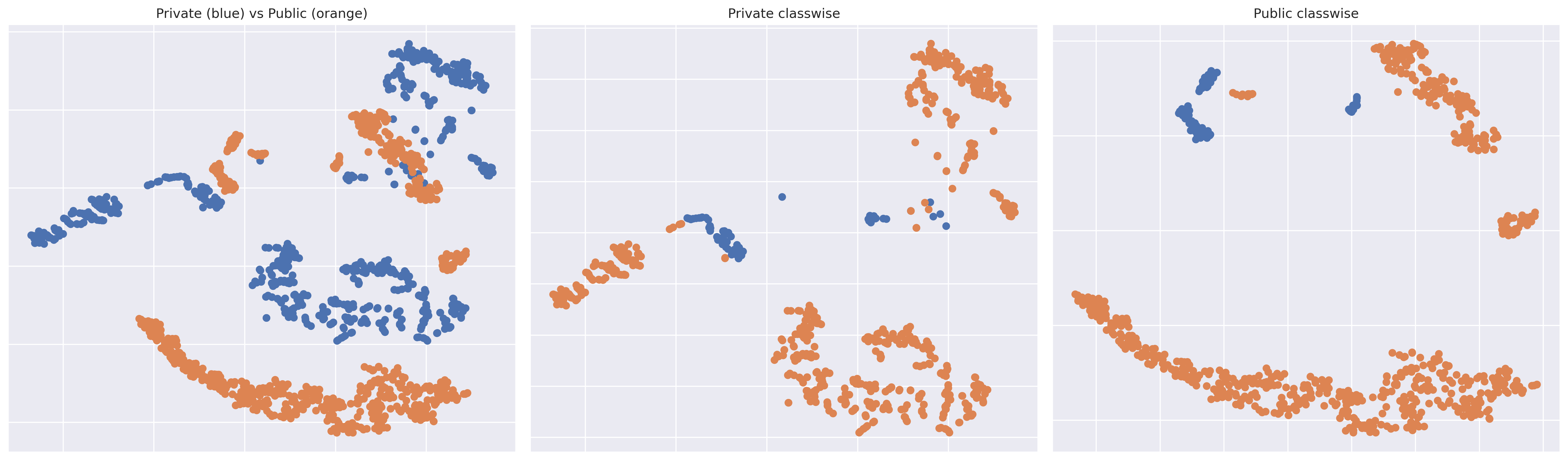}
\label{fig:wafer_tsne_dpwgan}
}
\subfloat[Wafer: GSWGAN~\cite{chen2020gs}]{
\includegraphics[width=.48\linewidth]{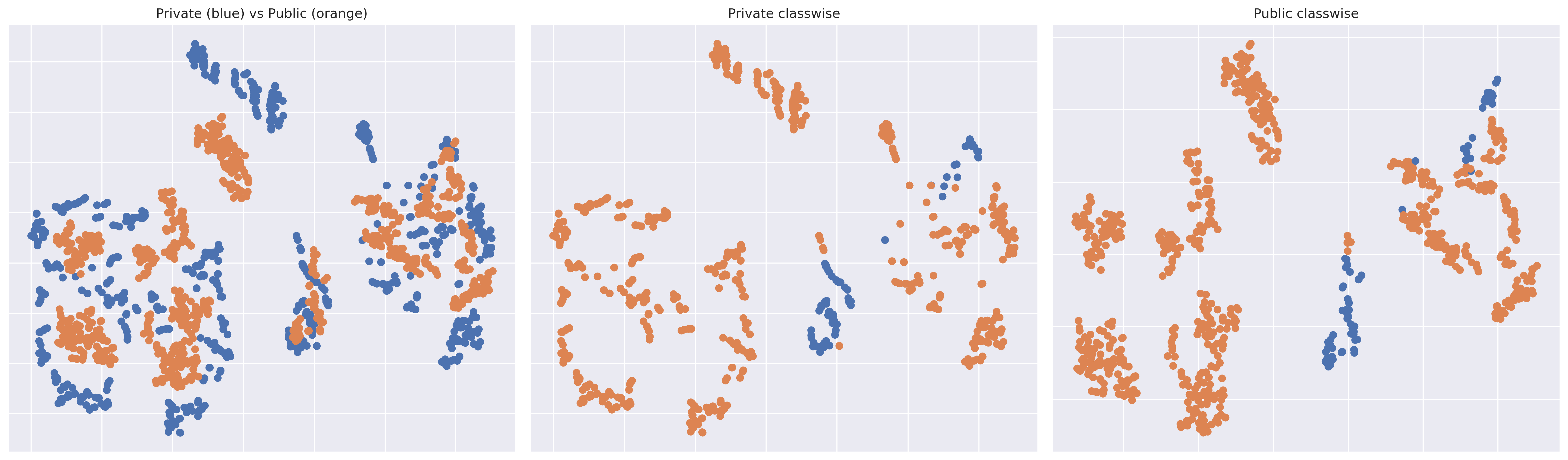}
\label{fig:wafer_tsne_gscwgan}
}
\hfil
\caption{T-SNE Visualization (1/2): Subfigures show datasets generated using DPWGAN~\cite{xie2018differentially} and GSWGAN~\cite{chen2020gs}. Plots within each subfigure: Left shows the difference between private and public train datasets. Middle: Class distribution of private train dataset. Right: Class distribution public train dataset.}
\label{fig:tsne}
\end{figure}

Understanding the actual dataset distribution is an important part, as in an ideal scenario, both datasets are indistinguishable. However, if the datasets are distinguishable, this does not mandatory result in a badly trained classifier. 
To understand the difference between the private and public data, T-SNE plots were created. The T-SNE approach transforms data into a two-dimensional space. The visualized plots show the difference between both sets. In Figure~\ref{fig:tsne} the results of four datasets are shown. The first column of each subfigure shows the private train data and the public train data. Ideally, the data overlaps, as this states that the private and generated (public) dataset are indistinguishable. The CharacterTrajectories dataset shows this behavior for DPWGAN~\cite{xie2018differentially} whereas, for the other three datasets, this is not the case. The second and third columns of the subfigure show the class distribution within the private and public train datasets. In the case of the CharacterTrajectories dataset, it is visible that for the DPWGAN~\cite{xie2018differentially} the class distribution shows a clear separation between the classes for both the private and public data (Figure~\ref{fig:char_tsne_dpgan}). Although the data shows a small offset, this did not hinder the performance by a large margin. Both models show a similar position of the 20 classes within the space. Furthermore, the results provide evidence that the dataset generated using GSWGAN~\cite{chen2020gs} shows a much better overlap compared to the private data. In addition, the class separation is much better compared to the DPWGAN~\cite{xie2018differentially}. The results provide evidence that the generated data for the CharacterTrajectories and FaceDetection dataset using the GSWGAN~\cite{chen2020gs} share the same distribution as the original data (Figure~\ref{fig:char_tsne_gscwgan} and Figure~\ref{fig:face_tsne_gscwgan}). Figure~\ref{fig:ecg_tsne_dpwgan} and Figure~\ref{fig:ecg_tsne_gscwgan} show the T-SNE plots for the ECG5000 dataset. This dataset consists of five classes. However, the first two classes cover 292 and 177 samples, leaving 31 samples for the remaining classes. This makes it impossible for the GANs to, correctly, learn all classes and results in generators that cover only the two main classes. In addition, Figure~\ref{fig:ecg_tsne_dpwgan} shows a large offset between the private and public data, which is reflected in the accuracy, as the public classifier cannot perform well on the private dataset. Furthermore, the two classes are not separated well. In contrast to that, Figure~\ref{fig:ecg_tsne_gscwgan} shows that the generated data of the GSWGAN~\cite{chen2020gs}, although it is different from the original distribution, separated the two classes well. In addition, this model was able to perform well, although the distribution shows some differences. For the accuracy results related to this figure, the reader is referred to Table~\ref{tab:comparison}. The last dataset is visualized in Figure~\ref{fig:wafer_tsne_dpwgan} and Figure~\ref{fig:wafer_tsne_gscwgan}. Similar to the previous dataset, the DPWGAN~\cite{xie2018differentially} produces a distribution with an offset, resulting in a bad performance. In contrast to that, the GSWGAN~\cite{chen2020gs} generated a dataset that shows a similar distribution within the T-SNE plot, resulting in high performance for the classifier trained on that dataset. 

One general finding across all datasets was that the DPWGAN~\cite{xie2018differentially} produces data that is less similar to the original data. This can be explained by the fact that differential privacy is applied to both the discriminator and generator, resulting in a worse performance of the discriminator. This is not the case for the GSWGAN~\cite{chen2020gs} as the privacy constraints are only applied to the generator. The stronger discriminator seems to improve the quality of the generated data.

\begin{figure}[!t]
\centering
\subfloat[AsphaltPavementType]{
\includegraphics[width=.31\linewidth]{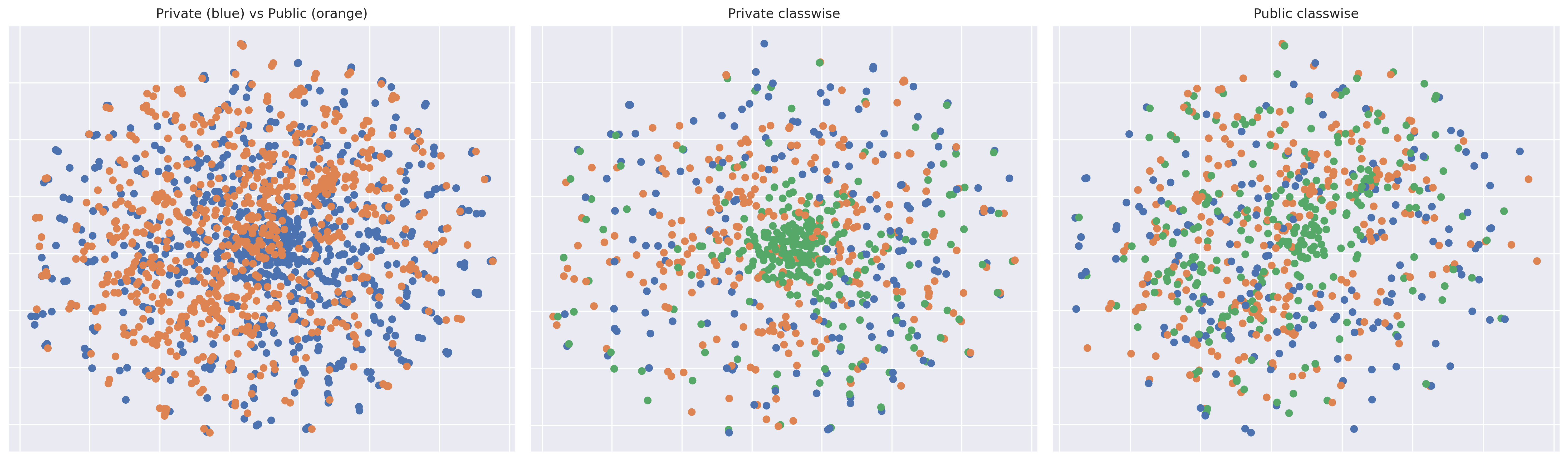}
\label{fig:pavements_tnse}
}
\subfloat[AsphaltRegularity]{
\includegraphics[width=.31\linewidth]{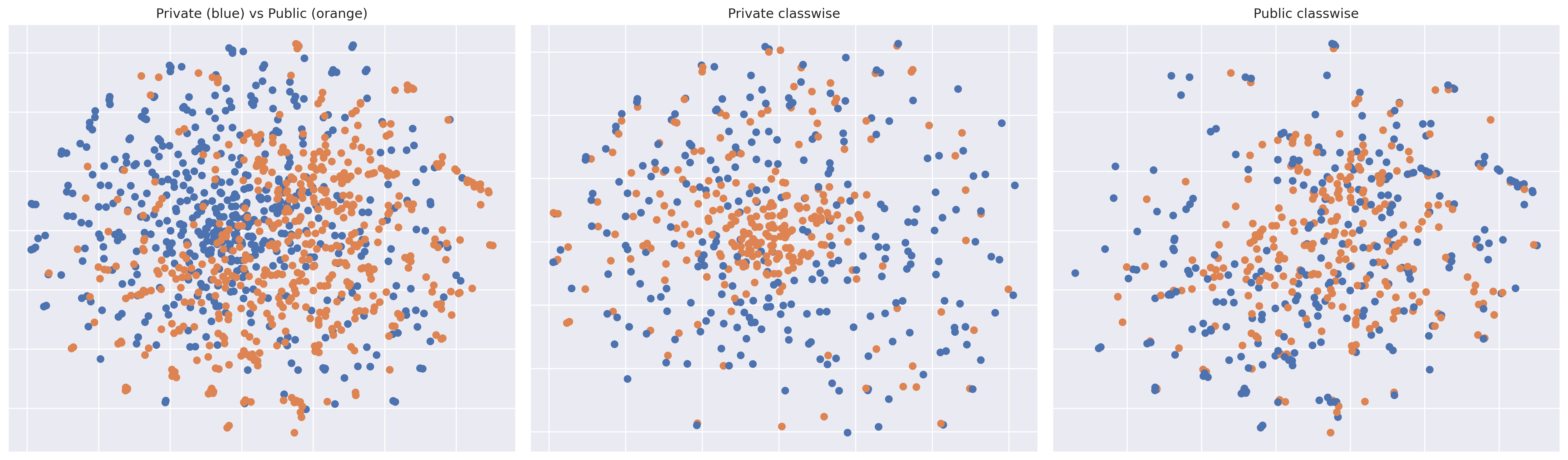}
\label{fig:regularity_tsne}
}
\subfloat[FordA]{
\includegraphics[width=.31\linewidth]{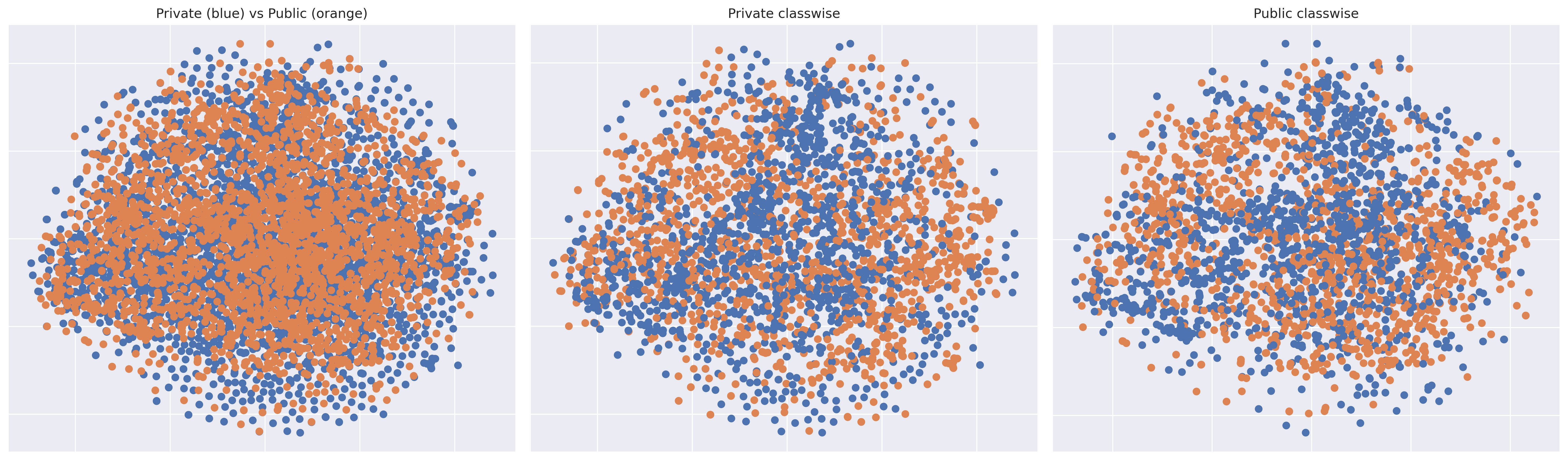}
\label{fig:forda_tsne}
}
\hfil
\subfloat[HandOutlines]{
\includegraphics[width=.31\linewidth]{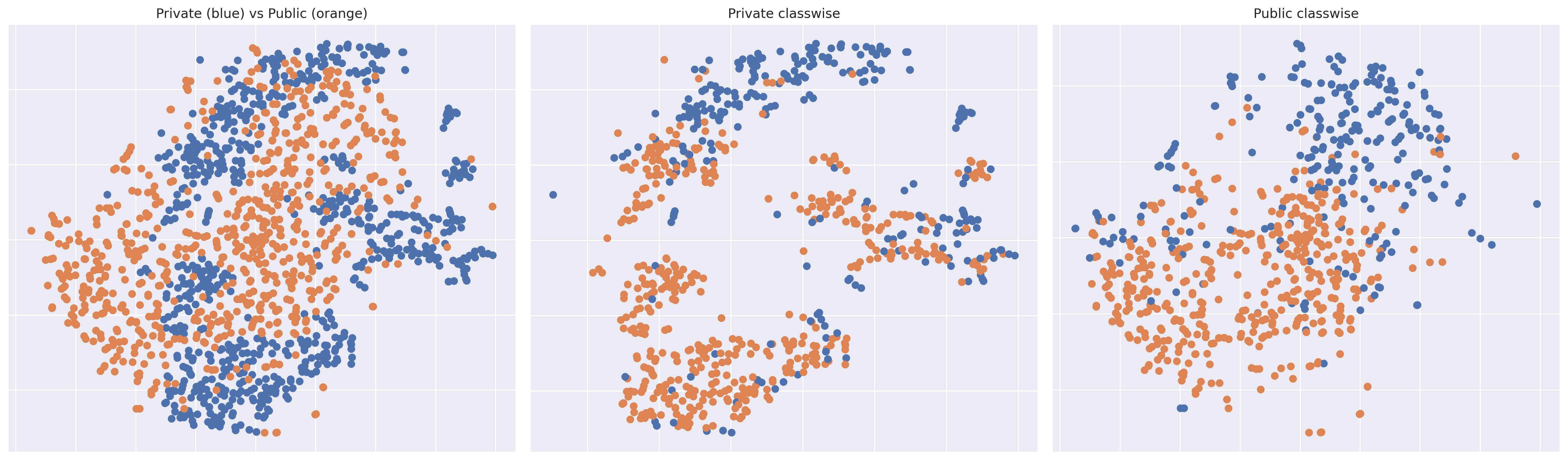}
\label{fig:outlines_tnse}
}
\subfloat[UWaveGestureLibraryAll]{
\includegraphics[width=.31\linewidth]{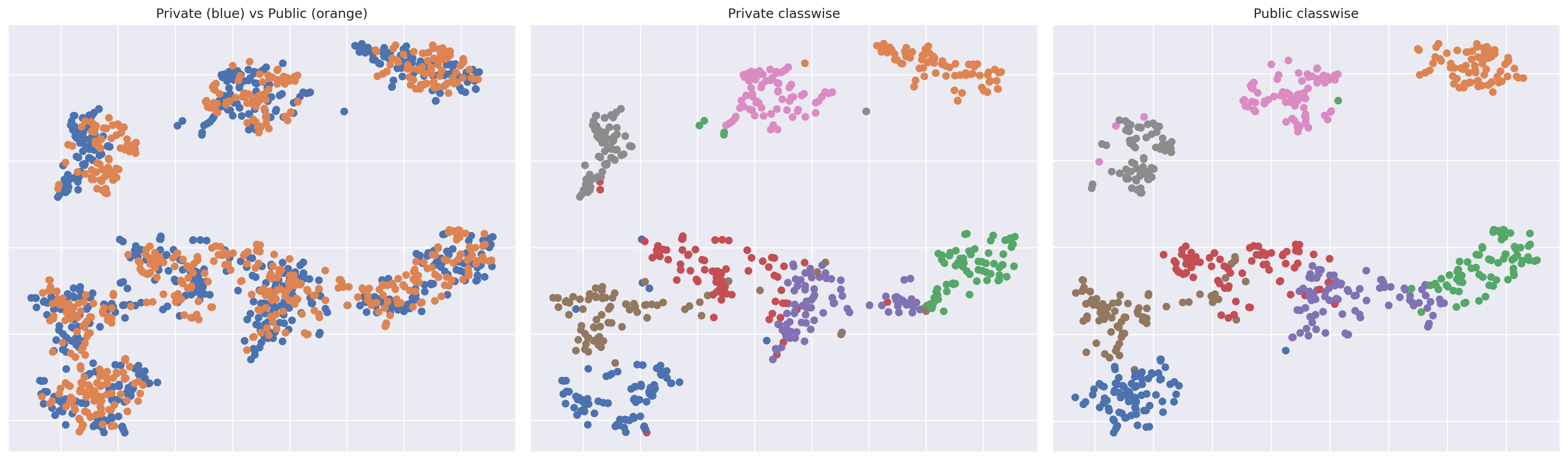}
\label{fig:gesture_tnse}
}
\hfil
\caption{T-SNE Visualization (2/2): Subfigures show the datasets generated using GSWGAN~\cite{chen2020gs}. Plots within each subfigure: Left shows the difference between private and public train datasets. Middle: Class distribution of private train dataset. Right: Class distribution public train dataset.}
\label{fig:tsne_extended}
\end{figure}

Figure~\ref{fig:tsne_extended} covers the remaining T-SNE plots for the GSWGAN~\cite{chen2020gs}. Except for the HandOutlines dataset, the dataset distributions of the private and the generated data show a high overlap. In addition, the class separation in the space is well for all the visualized datasets. Although the distribution of the HandOutlines dataset and the generated one do not perfectly fit, the results still show that the classifier trained on the generated dataset can infer the real data.

\subsection{Experiment 6: Dataset Visualization – Private vs. Generated (Public) Data}
\begin{figure}[!t]
\centering
\subfloat[CharacterTrajectories: Private data]{
\includegraphics[width=.47\linewidth]{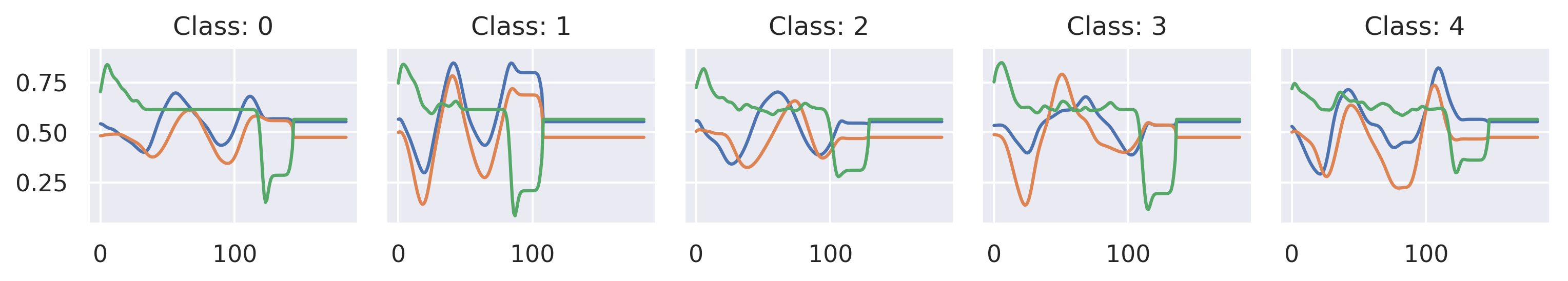}
\label{fig:char_real_data}
}
\subfloat[ECG5000: Private data]{
\includegraphics[width=.47\linewidth]{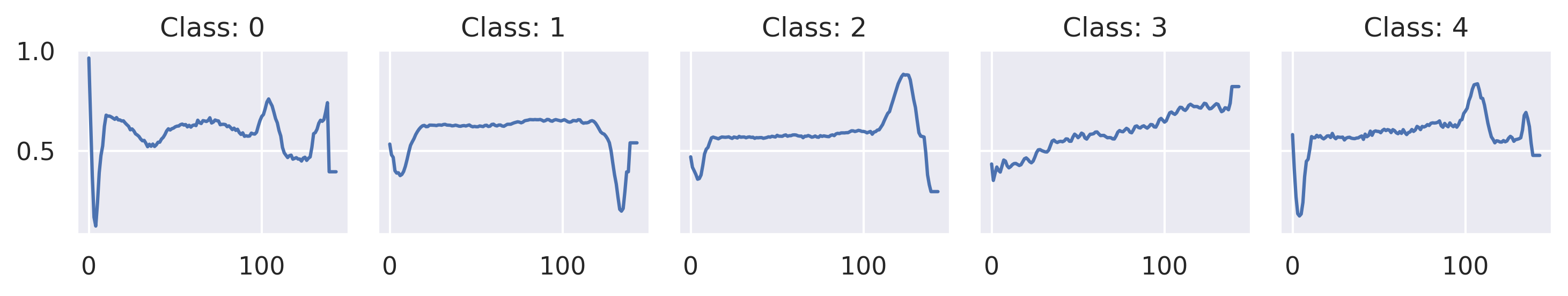}
\label{fig:ecg_real_data}
}
\hfil
\subfloat[CharacterTrajectories: Public data (GSWGAN-dense)]{
\includegraphics[width=.48\linewidth]{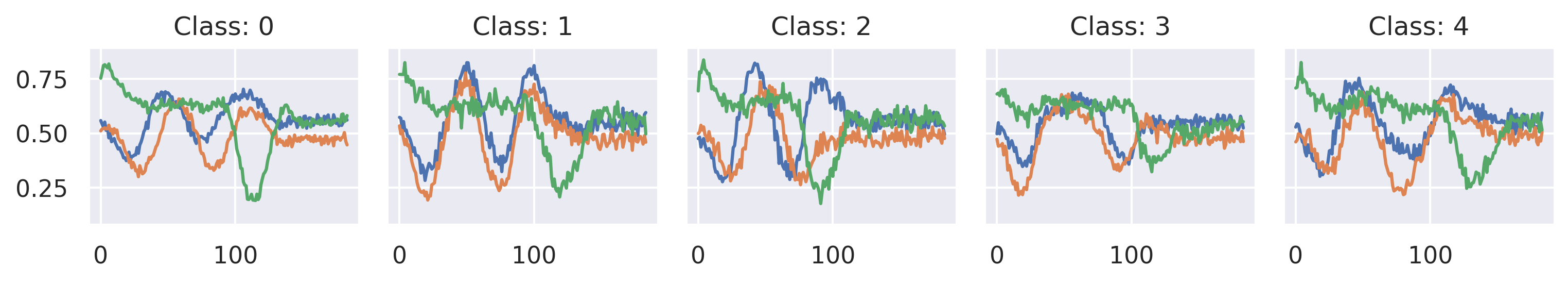}
\label{fig:char_dense_data}
}
\subfloat[ECG5000: Public data (GSWGAN-dense)]{
\includegraphics[width=.48\linewidth]{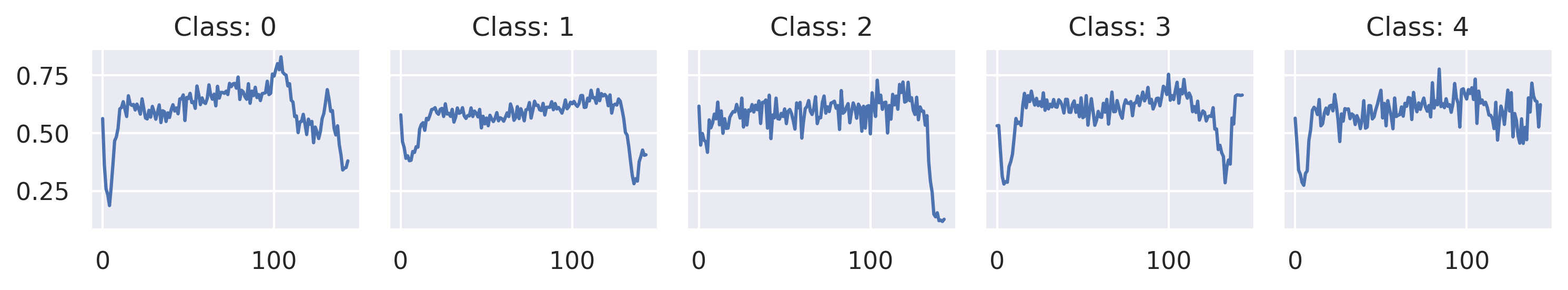}
\label{fig:ecg_dense_data}
}
\hfil
\subfloat[CharacterTrajectories: Public data (GSWGAN-conv)]{
\includegraphics[width=.47\linewidth]{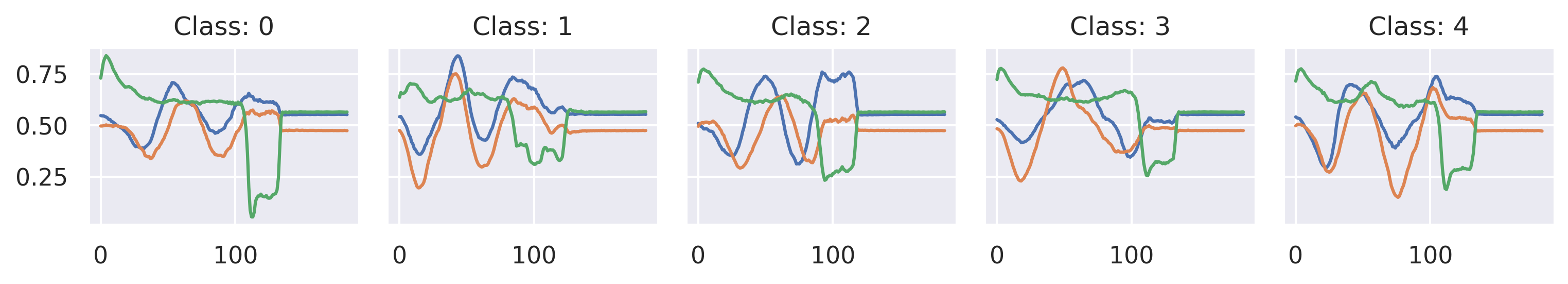}
\label{fig:char_conv_data}
}
\subfloat[ECG5000: Public data (GSWGAN-conv)]{
\includegraphics[width=.48\linewidth]{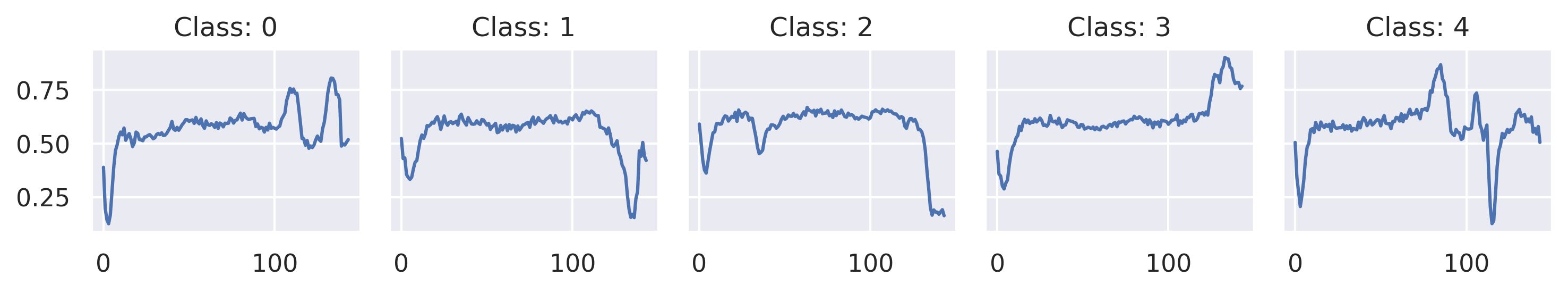}
\label{fig:ecg_conv_data}
}
\hfil
\subfloat[Wafer: Private data]{
\includegraphics[width=.32\linewidth]{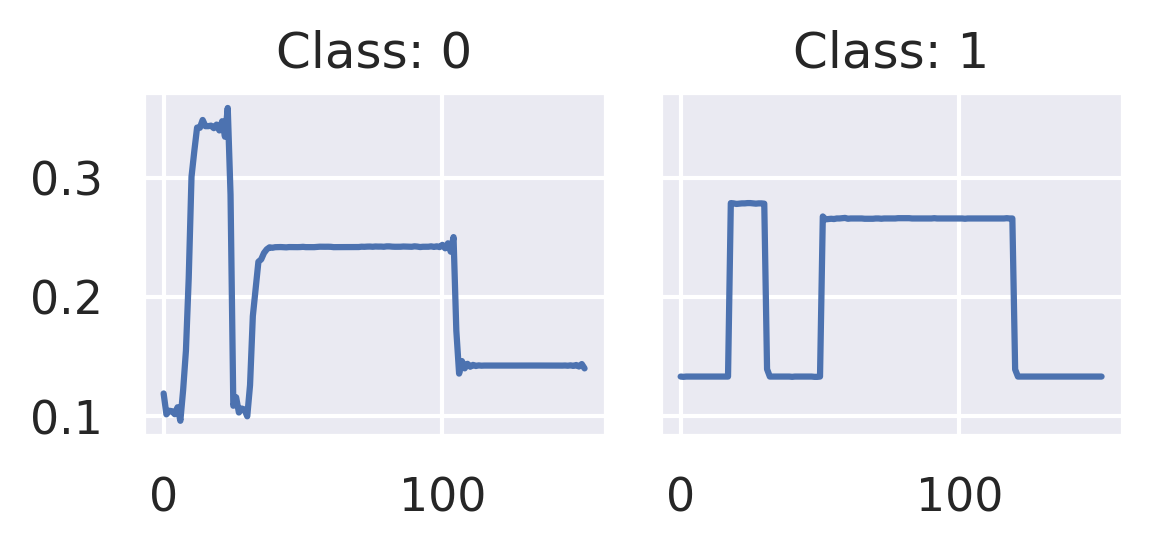}
\label{fig:wafer_real_data}
}
\subfloat[Wafer: Public data (GSWGAN-dense)]{
\includegraphics[width=.31\linewidth]{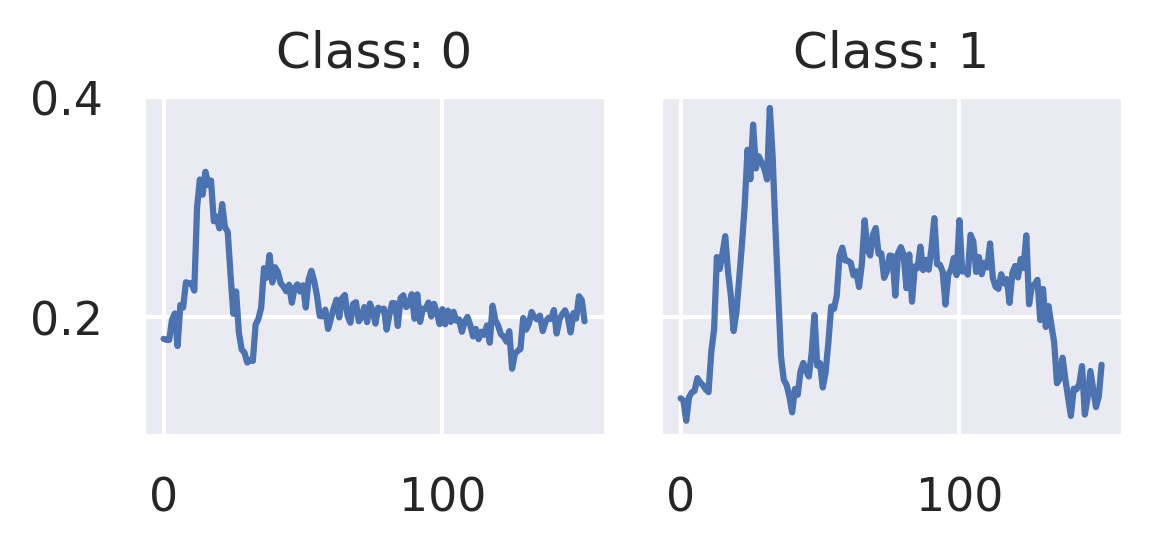}
\label{fig:wafer_dense_data}
}
\subfloat[Wafer: Public data (GSWGAN-conv)]{
\includegraphics[width=.31\linewidth]{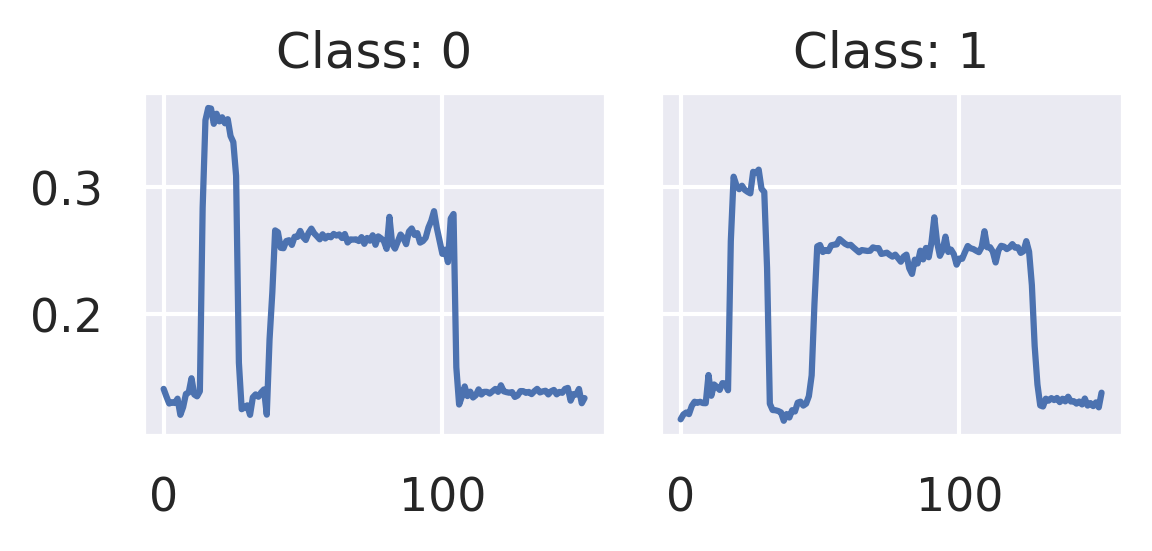}
\label{fig:wafer_conv_data}
}
\hfil
\caption{Dataset Visualization: Shows the private and the generated (public) data generated using GSWGAN-dense and GSWGAN-conv. The generated data of GSWGAN-dense shows a noisy behavior compared to the very realistic samples GSWGAN-conv~\cite{chen2020gs}.}
\label{fig:data}
\end{figure}

Breaking down the dataset using the T-SNE visualization shows that there is some discrepancy between the private and the generated data. However, understanding this discrepancy is key to understanding whether it is acceptable or not. Therefore, in Figure~\ref{fig:data} real data and generated data are visualized to highlight the main differences. To create the samples for GSWGAN-dense and GSWGAN-conv the corresponding generator was used. The results indicate that the data generated using the generator preserves the dataset characteristics and has a similar shape when compared to the original data. The samples of the FaceDetection dataset were excluded, as visualizing that numerous channels does not provide any insights. One finding is that using GSWGAN-conv results in much smoother samples compared to GSWGAN-dense. Especially, GSWGAN-dense shows worse performance on the Wafer dataset, although the shape of those samples is not very complex. On the other side, GSWGAN-conv was able to produce very smooth samples for the Wafer dataset. 

\begin{figure}[!t]
\centering
\subfloat[CharacterTrajectories]{
\includegraphics[width=.98\linewidth]{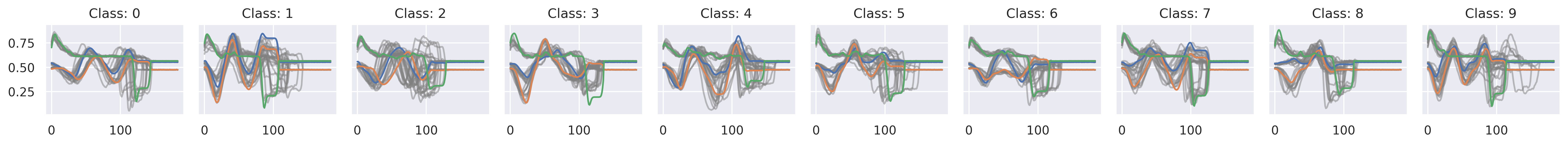}
\label{fig:char_data}
}
\hfil
\subfloat[ECG5000]{
\includegraphics[width=.69\linewidth]{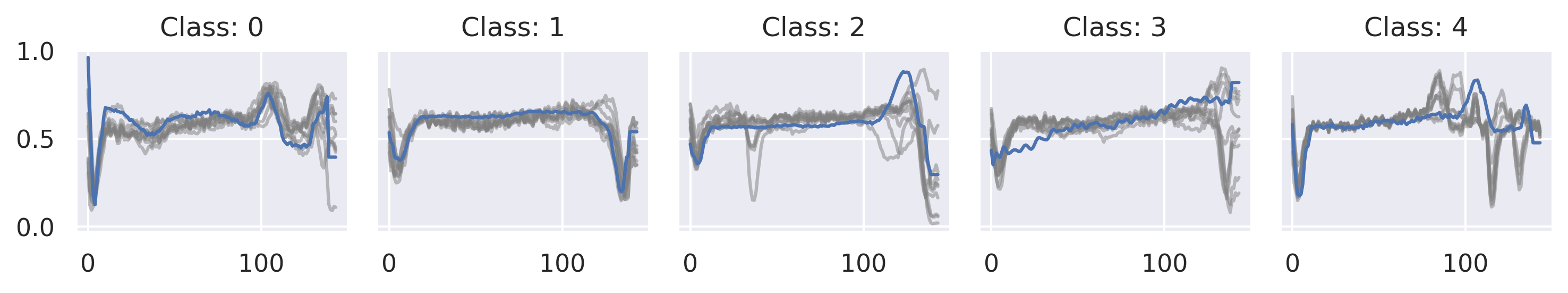}
\label{fig:ecg_data}
}
\subfloat[Wafer]{
\includegraphics[width=.27\linewidth]{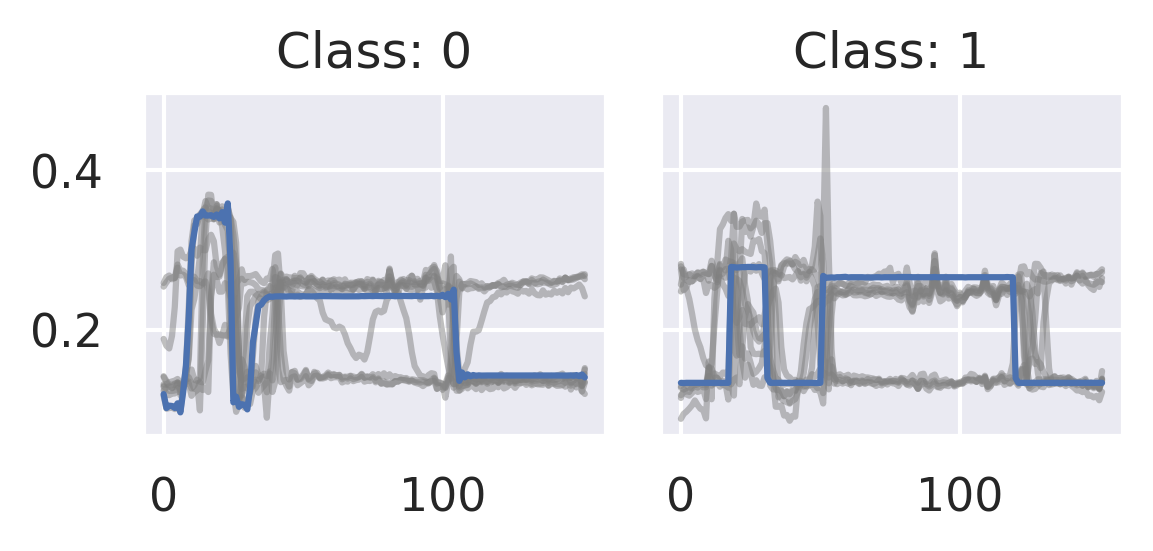}
\label{fig:wafer_data}
}
\hfil
\subfloat[AsphaltPavementType]{
\includegraphics[width=.41\linewidth]{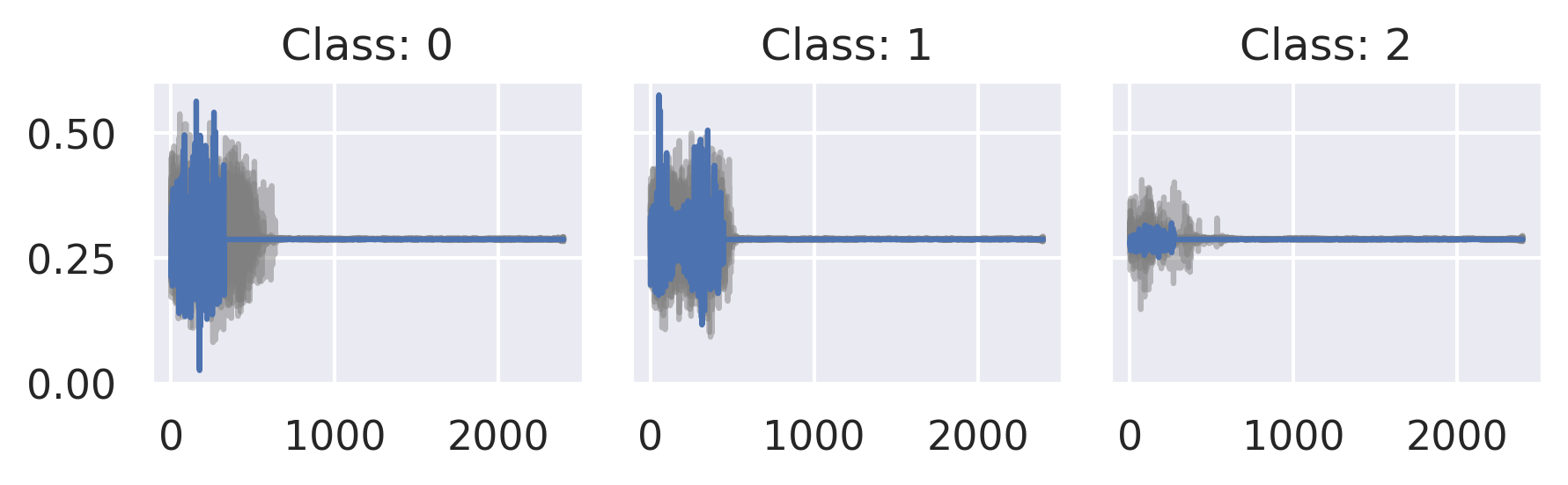}
\label{fig:pavements_data}
}
\subfloat[AsphaltRegularity]{
\includegraphics[width=.25\linewidth]{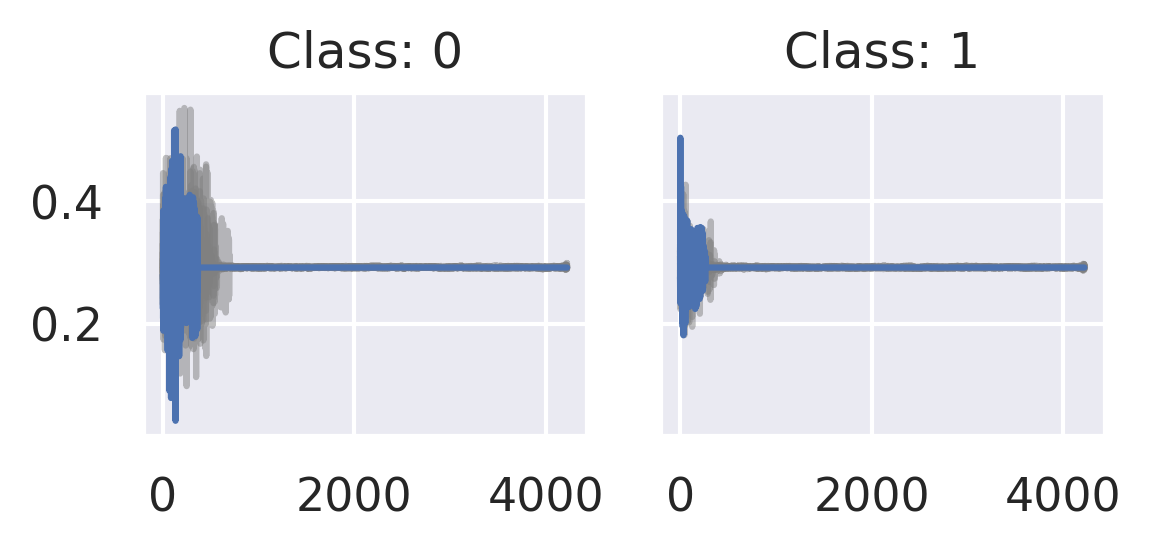}
\label{fig:regularity_data}
}
\subfloat[FordA]{
\includegraphics[width=.25\linewidth]{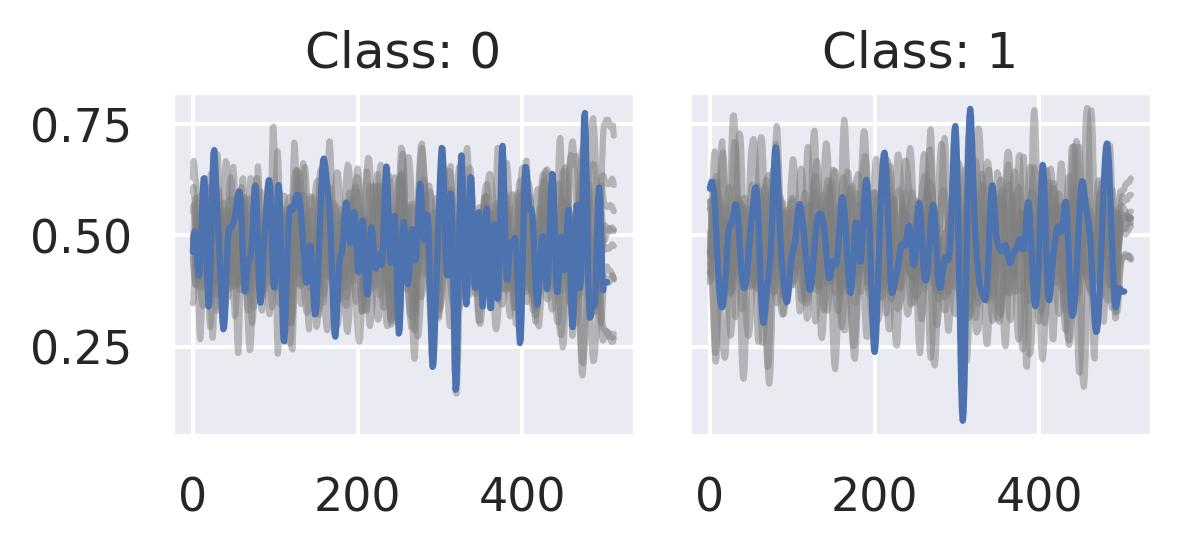}
\label{fig:forda_data}
}
\hfil
\subfloat[HandOutlines]{
\includegraphics[width=.25\linewidth]{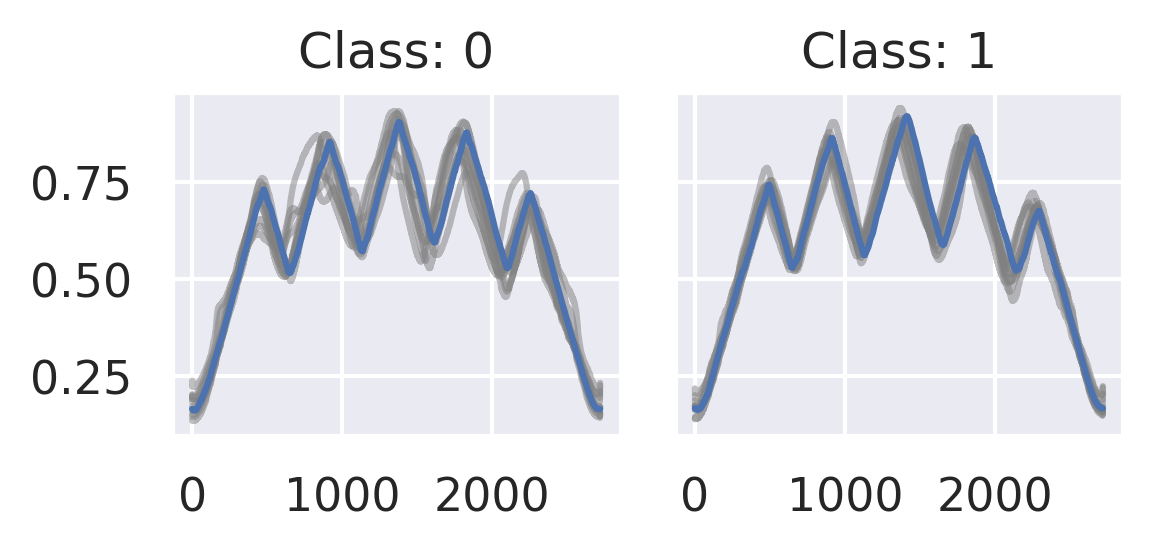}
\label{fig:outlines_data}
}
\subfloat[UWaveGestureLibraryAll]{
\includegraphics[width=.41\linewidth]{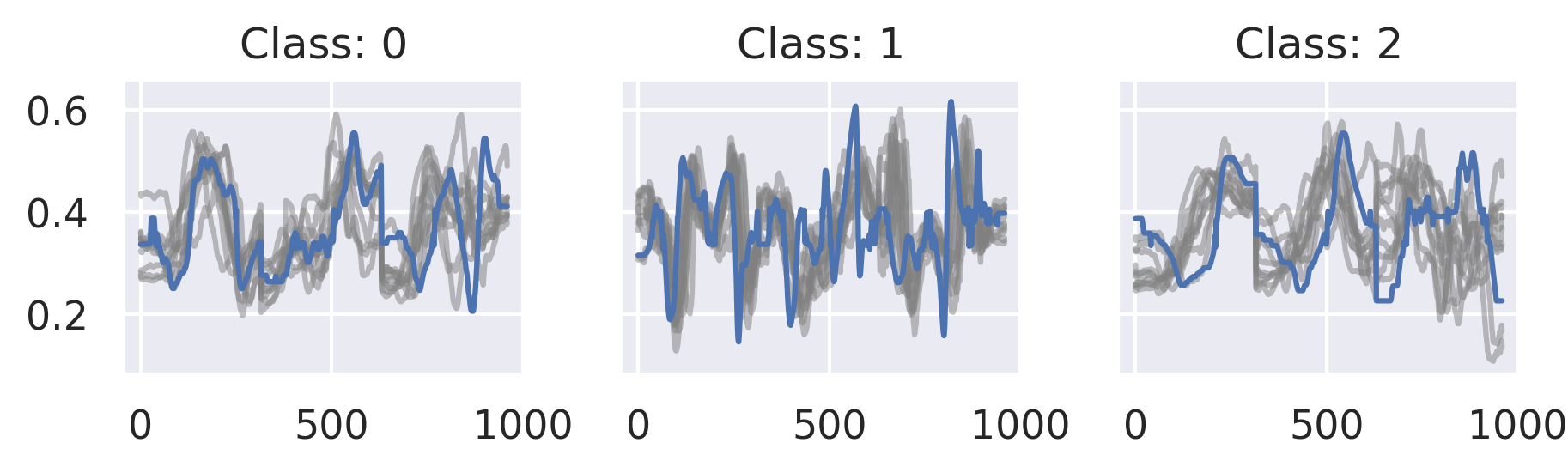}
\label{fig:gesture_data}
}
\hfil
\caption{Dataset Visualization: Shows datasets created using GSWGAN-conv~\cite{chen2020gs}. Grey lines correspond to multiple original data samples. Blue corresponds to the generated data sample.}
\label{fig:data_extended}
\end{figure}

In the second part of this experiment, a sample generated using GSWGAN-conv is plotted against the real data to show how it fits into the original dataset. GSWGAN-dense was excluded, as the previous experiments indicate that it provides lower-quality samples. It is visible that the general shape of the generated sample is similar. In a latter experiment, the dataset statistics are evaluated to show that it is not a direct copy of the existing data, as it is not feasible to infer that from this visualization. For privacy, it is important that the generator produce new samples and does not memorize the existing ones, as this yields data leaks. Figure~\ref{fig:data_extended} shows the real and generated samples for the datasets. The plots provide evidence that the generated samples are not identical to the real data, but still preserve the same overall shape related to the classes. Especially, for complex datasets such as the FordA dataset, it is necessary to have very smooth and high-quality samples as the anomaly detection task is fragile to peaks and small changes in the data. 

\subsection{Experiment 7: Dataset Analysis – Computing the Distance between Samples}
To provide evidence that the generated dataset does not consist of copies of the real data but rather shows new and different samples, the distances between samples were computed. The goal was to show that the samples are different but share the same distribution, which can not be stated by visual inspection. To analyze the differences and similarities between the private and the public datasets, the L2-norm was computed. Therefore, first, the distances within each dataset were computed. This helps to understand how diverse the data within the dataset is. However, to understand the connection between the datasets, the values were computed between the private and the public data. Therefore, the distance between every sample in the real dataset and the generated dataset was computed to identify changes in the minimal and maximal distance. Ideally, the distances should not change, highlighting that the data is neither a copy nor has the wrong distribution.

\begin{figure}[!t]
\centering
\subfloat[CharacterTrajectories]{
\includegraphics[width=.48\linewidth]{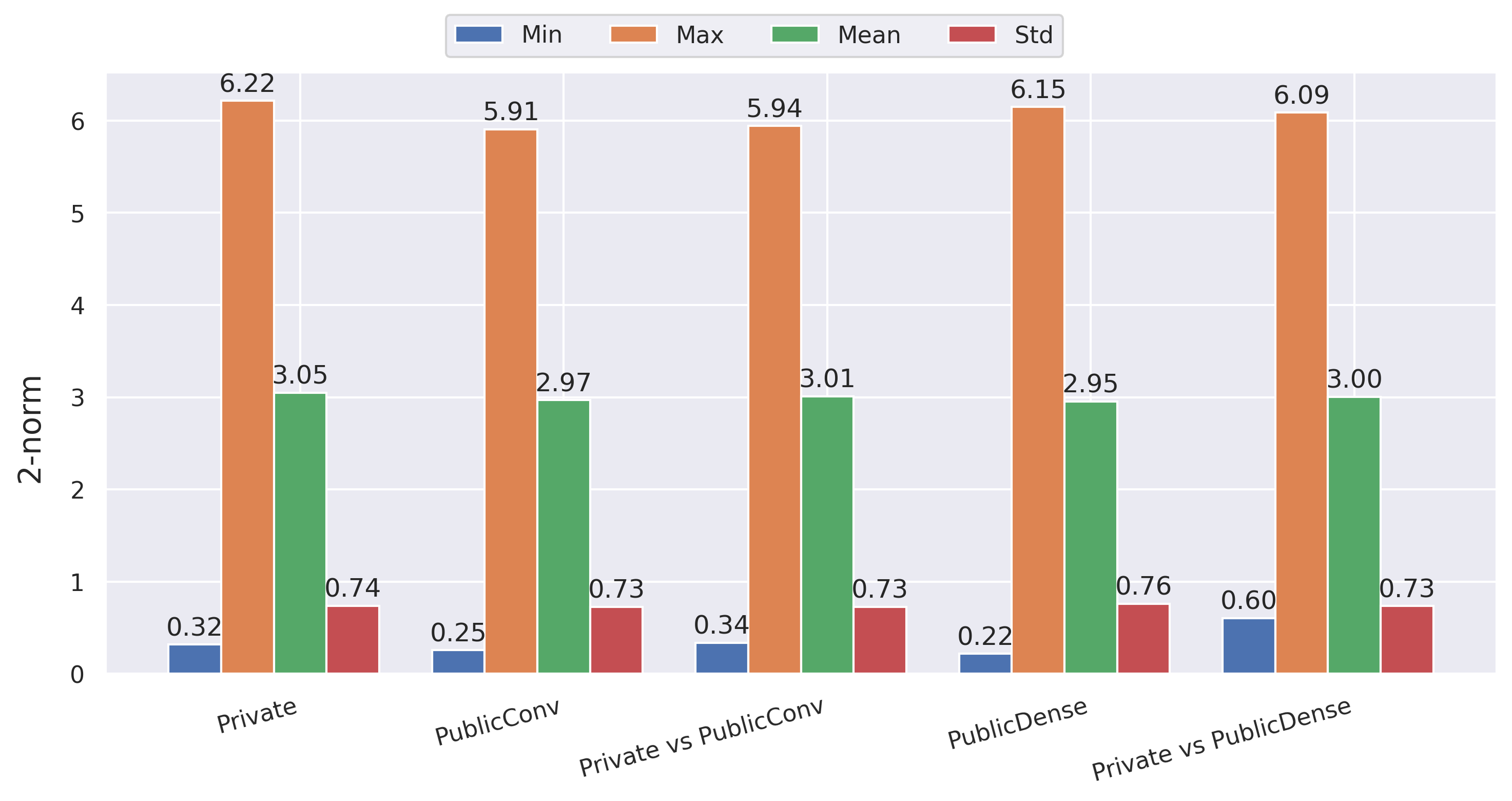}
\label{fig:char_data_metric}
}
\subfloat[ECG5000]{
\includegraphics[width=.48\linewidth]{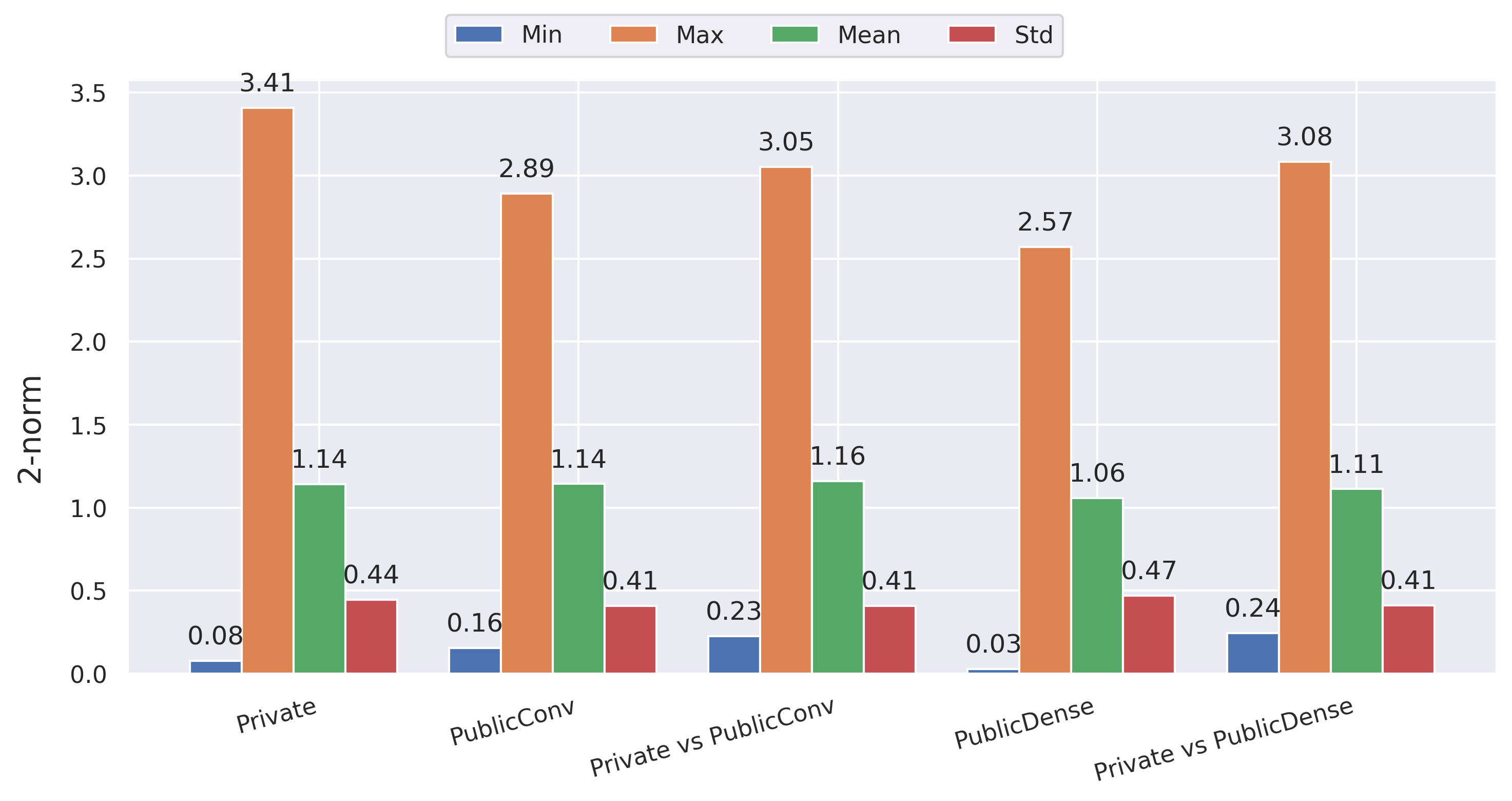}
\label{fig:ecg_data_metric}
}
\hfil
\subfloat[FaceDetection]{
\includegraphics[width=.48\linewidth]{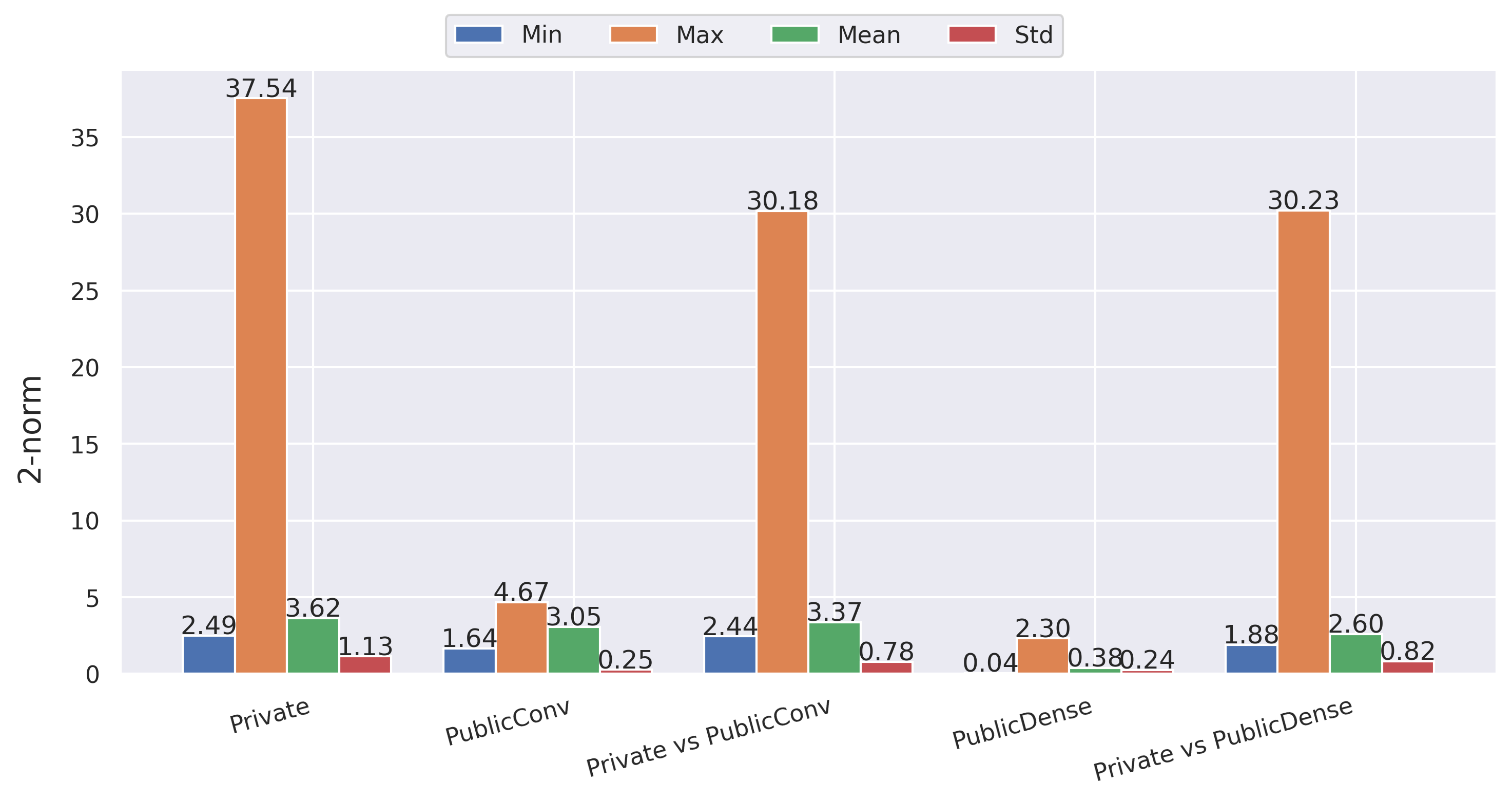}
\label{fig:face_data_metric}
}
\subfloat[Wafer]{
\includegraphics[width=.48\linewidth]{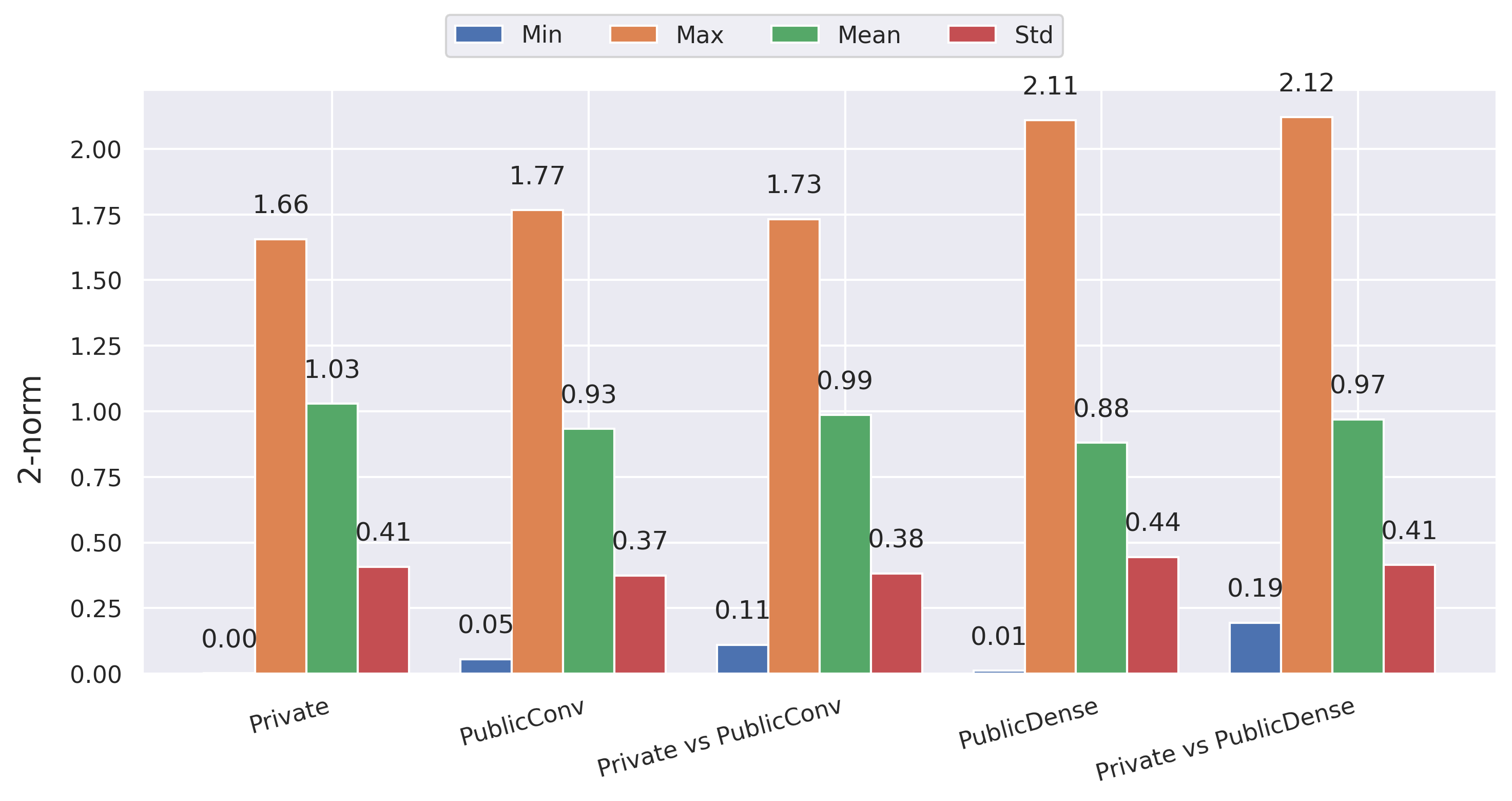}
\label{fig:wafer_data_metric}
}
\hfil
\caption{Dataset Distances (1/2): Shows the distance between the private and generated datasets using GSWGAN-dense and GSWGAN-conv and the samples within each dataset. To compare two datasets, the union of the samples was built and the L2-norm is used to compute the distance.}
\label{fig:data_metric}
\end{figure}

In Figure~\ref{fig:data_metric} the results are shown for four datasets which include the dense and convolutional setup. Especially, for the CharacterTrajectories the distances show that the distances within the private and the synthetic datasets are similar. The minimal distance value provides information about the clustering within the data, and ideally should be similar to the real private data. Otherwise, the synthetic samples could be centered around a single real data point. The maximum value provides insights about outliers. Comparing the values between the real and the generated dataset, the values are in the same range, highlighting that the data quality is good. A significantly lower minimum value would provide evidence that the generator copies existing data instead of creating new data. Accordingly, significantly higher maximum values correspond to outliers. The results show that except for the FaceDetection dataset, the distances provide evidence that the data quality is good. 

\begin{figure}[!t]
\centering
\subfloat[AsphaltPavementType]{
\includegraphics[width=.31\linewidth]{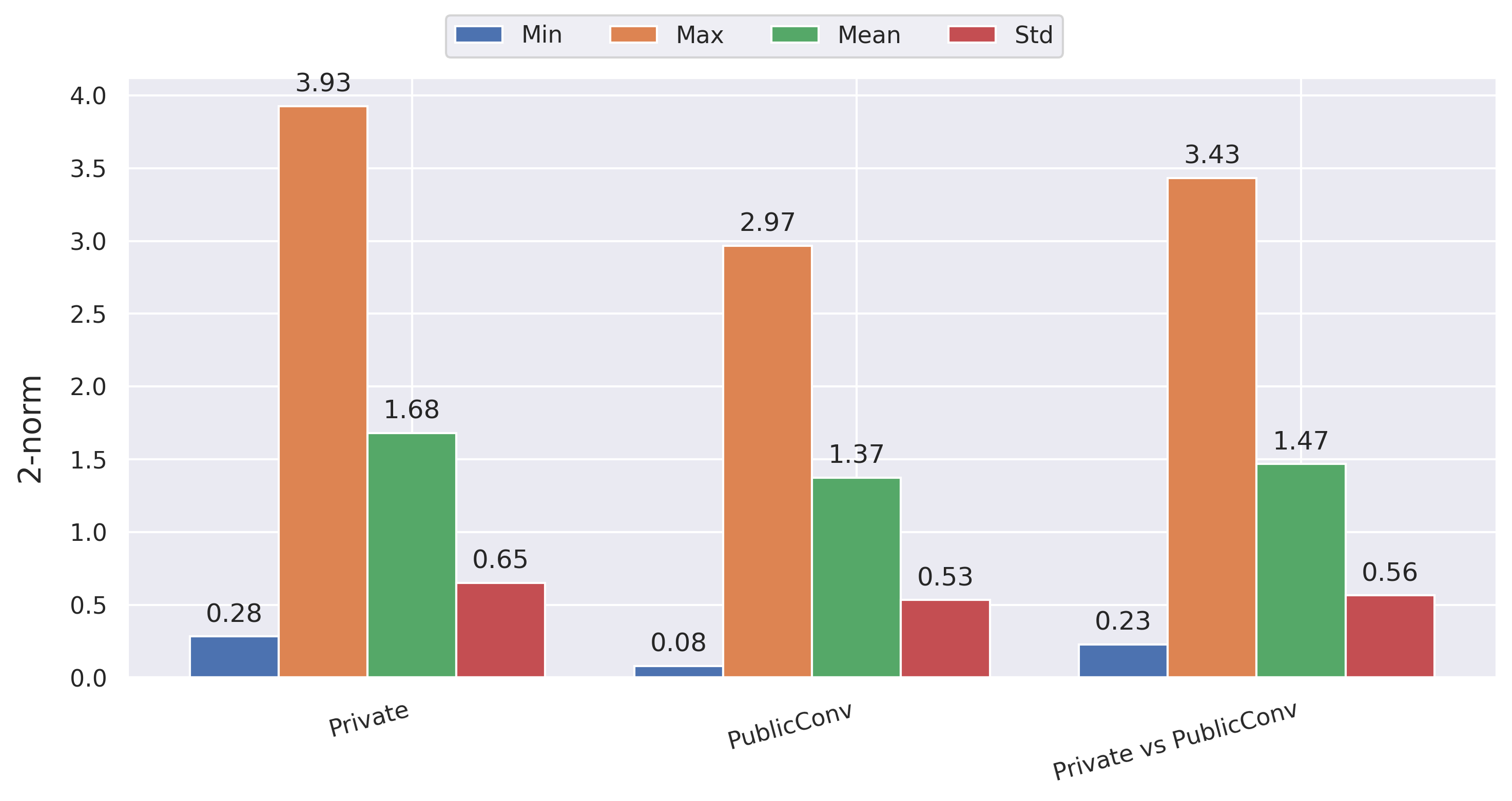}
\label{fig:pavements_distance}
}
\subfloat[AsphaltRegularity]{
\includegraphics[width=.31\linewidth]{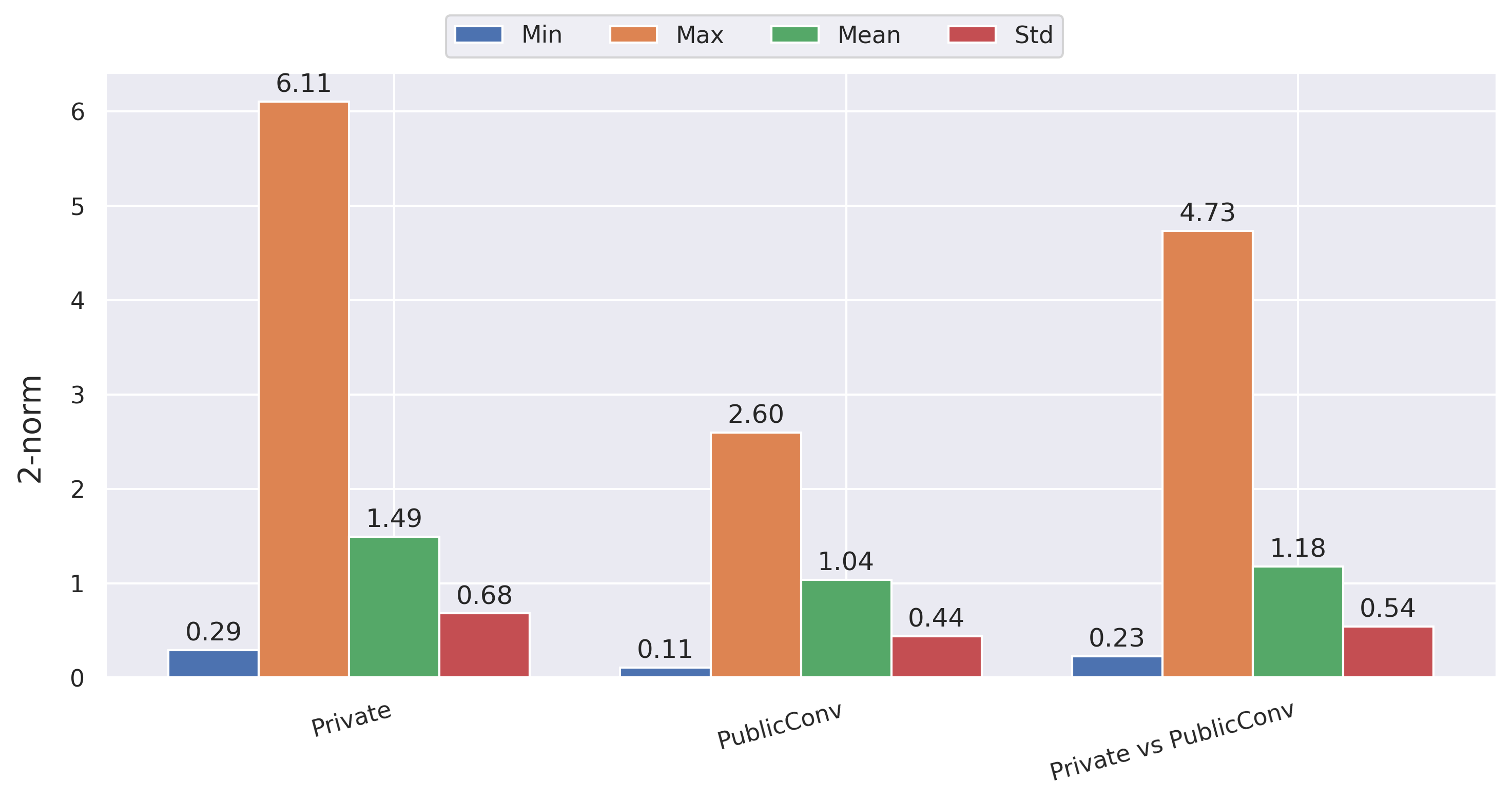}
\label{fig:regularity_distance}
}
\subfloat[FordA]{
\includegraphics[width=.31\linewidth]{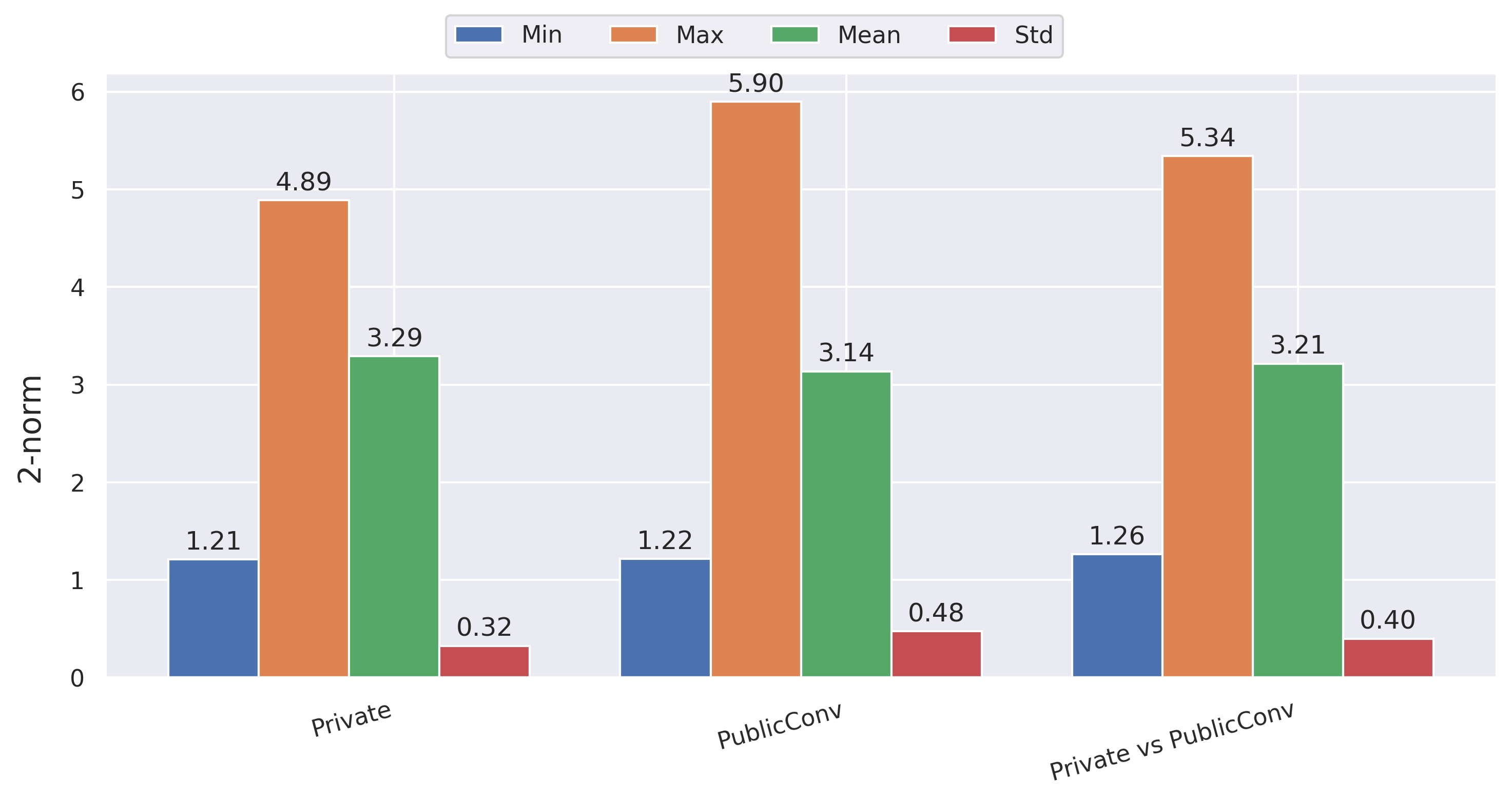}
\label{fig:forda_distance}
}
\hfil
\subfloat[HandOutlines]{
\includegraphics[width=.31\linewidth]{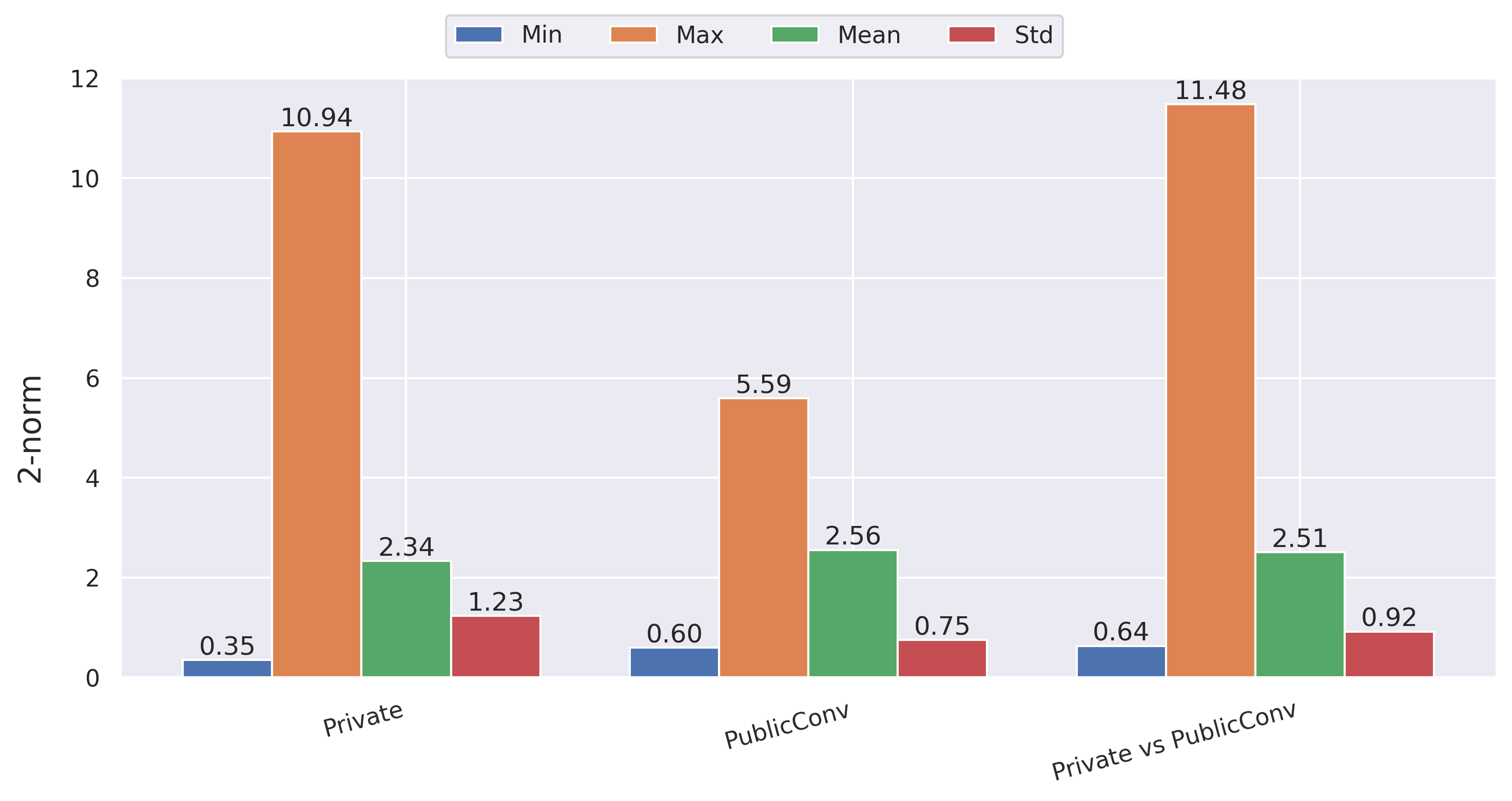}
\label{fig:outlines_distance}
}
\hfil
\subfloat[UWaveGesture]{
\includegraphics[width=.31\linewidth]{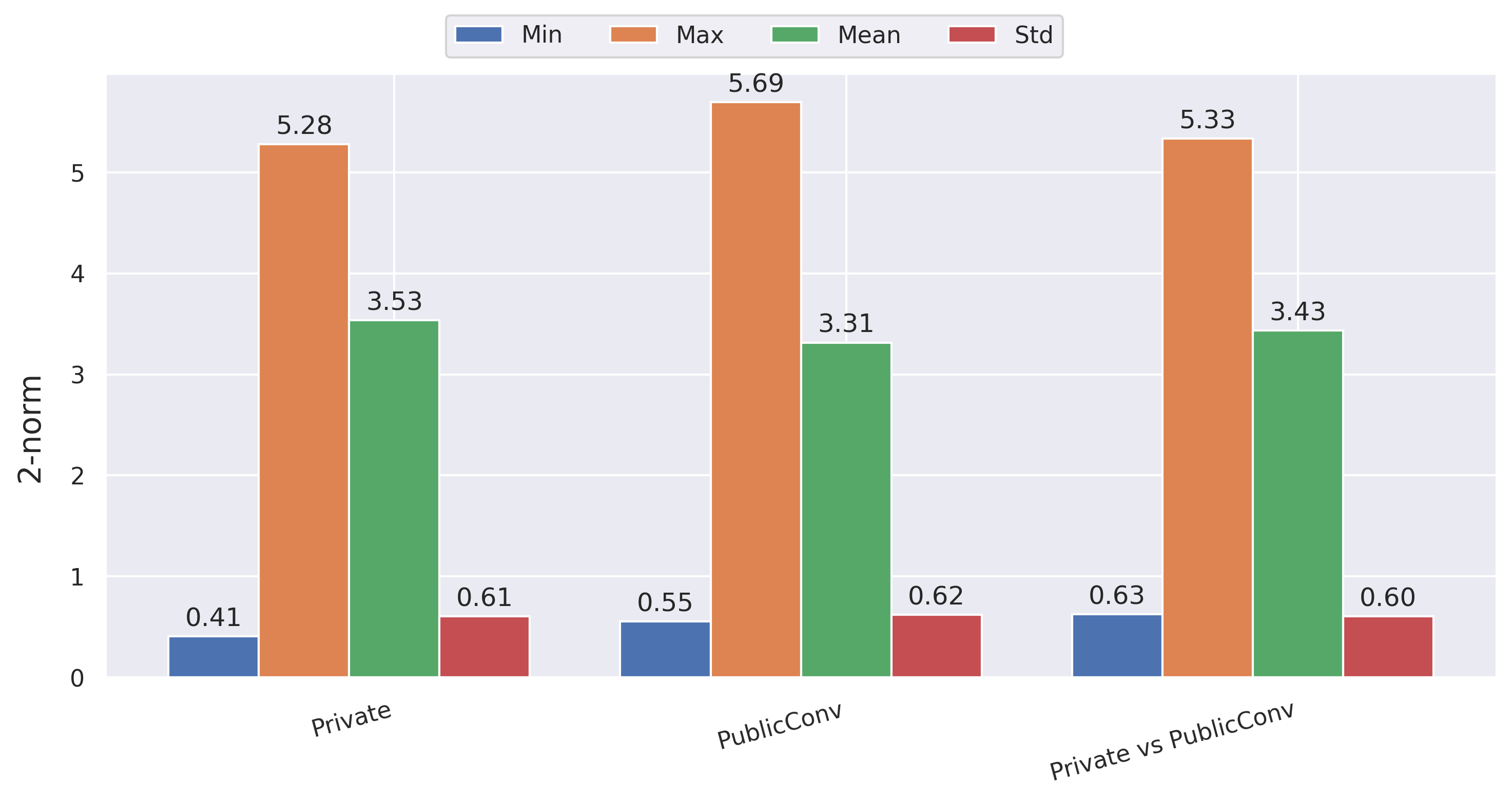}
\label{fig:gesture_distance}
}
\hfil
\caption{Dataset Distances (2/2): Shows the distance between the private and generated datasets for GSWGAN-conv and the samples within each dataset. To compare two datasets, the union of the samples was built and the L2-norm is used to compute the distance.}
\label{fig:data_metric_extended}
\end{figure}

However, in Figure~\ref{fig:face_data_metric} the maximum distance within the synthetic datasets is much lower compared to its real data counterpart. This suggests that the generated data does not show a high variance. Furthermore, the mean distance for GSWGAN-dense is very low, which suggests that most of the data is very similar, and the synthetic data does not cover the complete distribution. In contrast to that, GSWGAN-conv shows a similar mean distance as the real data. The comparison between the real and synthetic data provides further evidence that the generated data of the GSWGAN-conv generator successfully generates data that preserves a certain distance from the real data. For the remaining datasets in Figure~\ref{fig:data_metric_extended}, similar results can be observed. There are only small changes in the distance values, highlighting that the method successfully produces new data samples within the same space. For these datasets, only the convolutional setups are visualized, as these produced much better results.

\section{Discussion}
GSWGAN~\cite{chen2020gs} has shown a good performance across all the experiments. The comparison of the privacy approaches has shown that while a differential private classifier might apply to some cases, it limits not only the performance but is less stable than the GSWGAN~\cite{chen2020gs}. Furthermore, it can have impact methods applied on top of the classifier, such as interpretability methods. Another finding was that the DPWGAN~\cite{xie2018differentially} has shown inferior performance for all experiments compared to the  GSWGAN~\cite{chen2020gs}. The lower performance can be explained by the noise added to the complete architecture, whereas the GSWGAN~\cite{chen2020gs} only produces a private generator. To produce this private generator, the privacy of the discriminator is not mandatory, and therefore the stronger discriminator makes the network more stable. Furthermore, the experiments on the architecture strongly evidence that GSWGAN-conv outperforms GSWGAN-dense in the context of time series. The convolutional generator produced smoother samples and achieved higher accuracies. Especially, for anomaly detection tasks, it is important to use the convolutional setup for smoother samples. Furthermore, depending on the dataset, the successful application of GSWGAN~\cite{chen2020gs} requires an architecture search to find a suitable network. The T-SNE visualization has shown similar findings concerning the better convergence of GSWGAN~\cite{chen2020gs} compared to DPWGAN~\cite{xie2018differentially}. The distribution of the generated datasets projected into a two-dimensional space is very similar to the original dataset. Finally, the dataset visualization and dataset statistics provide evidence that the generated samples are smooth and similar to the real data but still different enough from the real data in a way that the L2-norm between the real and generated data are similar. 

The overall conclusion is that it is possible to apply GSWGAN~\cite{chen2020gs} on time series data to create public data from private datasets and enable the use of this data without additional constraints. This is not possible with the differential private classifier, and it offers giant potential as data can be shared securely and the outcome of the public model can be directly applied to the private data as the distribution of both sets is equal. This is especially of interest for private domains such as the health care and financial domain in which the sharing of the data is not possible, but explanations are required.

\section{Conclusion}
This paper benchmarked DPWGAN and GSWGAN in the context of time series classification. The results provide evidence that the gradient-sanitized approach is superior to the traditional DPWGAN. GSWGAN was able to achieve more stable performance across the datasets with the same amount of privacy. In addition, training a classifier directly, on private data does not provide better results across all datasets, highlighting that the generation of public data makes it possible to use any classifier without the limitation of differential privacy. The original GSWGAN uses a fixed number of iterations to train. However, the results indicate that the quality of the generator is not monotonic increasing and early stopping criteria such as the FID score resulted in superior performance for all datasets. Furthermore, the experiments indicate that GSWGAN-conv provides much smoother samples compared to GSWGAN-dense. In addition, the experiments provided evidence that an architecture search has a significant impact on the performance of the generative model. Concerning privacy, GSWGAN further shows better results when increasing the noise multiplier to achieve better privacy values. Finally, visual evidence for the correct dataset creation was provided and the distances within and between the real (private) and generated (public) data were computed, which validated the correctness of GSWGAN in the data generation process. The generator was able to provide high-quality data that share the same distribution as the real data.

\setlength{\bibsep}{.8em}
\bibliography{bibliography}

\end{document}